\documentclass[journal]{IEEEtran}
\usepackage{ifpdf}
  \usepackage{subfigure, enumerate}
  \usepackage{comment}  
  \usepackage{algorithm}
  \usepackage{algpseudocode}
  \usepackage{changes}

\usepackage[space]{cite}

\ifCLASSINFOpdf
   \usepackage[pdftex]{graphicx}
\else
\fi
\usepackage{amsmath}
\usepackage{array}
\usepackage{fixltx2e}

\usepackage{stfloats}
\usepackage{url}

\begin{document}
%
\title{Trajectory-based Scene Understanding using Dirichlet Process Mixture Model}
%
%
%

\author{Kelathodi~Kumaran~Santhosh,~\IEEEmembership{Student Member,~IEEE,}
        Debi~Prosad~Dogra,~\IEEEmembership{Member,~IEEE,}
        Partha~Pratim~Roy~\IEEEmembership{Member,~IEEE,}       
        and~Bidyut~Baran~Chaudhuri,~\IEEEmembership{Life~Fellow,~IEEE}
\thanks{K.~K.~Santhosh and D.~P.~Dogra are with School of Electrical Sciences, Indian Institute of Technology Bhubaneswar, Odisha,
India e-mail: (sk47@iitbbs.ac.in, dpdogra@iitbbs.ac.in).}
\thanks{P.~P.~Roy is with the Department of Computer Science and Engineering, Indian Institute of Technology, Roorkee, India. e-mail:(proy.fcs@iitr.ac.in).}
\thanks{B.~B.~Chaudhuri is a Professor, Vice chancellor, Techno India University, Saltlake city, Sector V, Kolkata, India. He is also a honorary visiting Professor at Computer Vision and Pattern Recognition Unit, Indian Statistical Institute, Kolkata, India. e-mail:(bbcisical@gmail.com)}
}

%
%

\markboth{PRE-PRINT}%
{Shell \MakeLowercase{\textit{et al.}}: Bare Demo of IEEEtran.cls for IEEE Journals}
%



\maketitle

\begin{abstract}
Appropriate modeling of a surveillance scene is essential for detection of                                                                                                                                                                                                                                        anomalies in road traffic. Learning usual paths can provide valuable insight into road traffic conditions and thus can help in identifying unusual routes taken by commuters/vehicles. If usual traffic paths are learned in a nonparametric way, manual interventions in road marking road can be avoided.  In this paper, we  propose an unsupervised and nonparametric method to learn frequently used paths from the tracks of moving objects in $\Theta(kn)$ time, where $k$ denotes the number of paths and $n$ represents the number of tracks. In the proposed method, temporal dependencies of the moving objects are considered to make the clustering meaningful using Temporally Incremental Gravity Model (TIGM). In addition, the distance-based scene learning makes it intuitive to estimate the model parameters. Further, we have extended TIGM hierarchically as Dynamically Evolving Model (DEM) to represent notable traffic dynamics of a scene. Experimental validation reveals that the proposed method can learn a scene quickly without prior knowledge about the number of paths ($k$). We have compared the results with various  state-of-the-art methods. We have also highlighted the advantages of the proposed method over existing techniques popularly used for designing traffic monitoring applications.  It can be used for administrative decision making to control traffic at junctions or crowded places and generate alarm signals, if necessary. 
\end{abstract}
\begin{IEEEkeywords}
Dirichlet process mixture model, Gibbs sampling, Bayesian inference, Incremental trajectory clustering, Intelligent transportation system, Nonparametric model, Unsupervised learning.
\end{IEEEkeywords}

%
\IEEEpeerreviewmaketitle

\section{Introduction}
%
%
%
%
\vspace{-1mm}
\IEEEPARstart{I}{n} the event of increasing volume of traffic on roads, it is difficult for human operators to monitor and understand the behavior of moving objects in real-time. Developing computer vision-guided  Intelligent transportation systems (ITS) can help the administration to learn the traffic behavior efficiently.  

If the parameters of the model representing the scene can be learned,  traffic behavior analysis is possible with such knowledge~\cite{2012_AA_Sodemann, 2015_B_Tian}. With the widespread use of surveillance cameras, traffic video feeds are becoming useful sources to understand the scene. Moreover, due to rapid advances in machine learning algorithms, it is now possible to infer hidden patterns using unsupervised learning techniques~\cite{cancela2014unsupervised, IncrementalDPMM}. A large segment of the existing unsupervised learning techniques requires the number of possible patterns to be specified apriori~\cite{2015_B_Tian}. However,  nonparametric modeling of data is one of the key features when the number of hidden patterns are not known in advance.  

Learning normal trajectory patterns is a key to understand the traffic behavior. Traffic situations such as congestion, illegal stoppages of vehicles, reverse driving, crashes, lane breaking, illegal U-turns, pedestrian movement on road can be categorized as anomaly~\cite{2012_AA_Sodemann} depending on the context and application. It is possible to distinguish between normal and abnormal patterns if the parameters of the model representing normal events can be learned. Moreover, traffic behavior may change over time during the course of the day. Hence the model representing the scene dynamics needs to be adaptive.

Ideally, a traffic analysis framework is expected to learn the parameters of the model in an unsupervised and nonparametric way. Therefore, such a framework has been proposed. It is based on a modified Dirichlet Process Mixture Model (DPMM). Sequential dependencies among the trajectories have been utilized in the proposed method that are typically ignored by the traditional clustering schemes~\cite{comaniciu2002mean,ester1996density}.  This has helped to achieve better performance without compromising the accuracy. The proposed model referred to as TIGM is extended in the time domain as DEM to understand the dynamics of different traffic paths. This can help the authorities to understand how the traffic changes over time in order to devise better transportation strategies. Knowledge about the traffic volume can also help the authorities to do efficient monitoring, congestion management, traffic flow analysis, traffic signal management, traffic planning for road construction, etc.
\vspace{-3mm}
\subsection{Related Work}
Trajectory learning and classification are two important components of video analytic methods. Trajectory-based methods have been used for flow analysis~\cite{4}, classification~\cite{5, 9}, activity recognition~\cite{6}, interaction analysis~\cite{7}, abnormal object detection~\cite{8}, and  object classification and tracking~\cite{10}. The aforementioned research work use supervised approaches, where labeled data are required. Unsupervised approaches have been used for vehicle behavior understanding\cite{11}, trajectory classification~\cite{12}, trajectory clustering\cite{13,14, IncrementalDPMM, cancela2014unsupervised}, etc. These methods employ unlabeled data to cluster similar trajectories and use clustered data to train the models for classification. Incremental clustering has been used in  trajectory modeling~\cite{IncrementalDPMM} and anomaly detection~\cite{16}, where the  data is processed sequentially.

    Typically, trajectory learning methods use extracted trajectory features for clustering. Mean shift is used in~\cite{13} and~\cite{21} with  Discrete Wavelet Transform (DWT) coefficients and multi-feature vector, respectively. Particle swarm optimization-based clustering of~\cite{izakian2016automated} with Dynamic Time Wrapping (DTW) uses all the trajectory points. The work proposed in \cite{wang2011trajectory} performs trajectory clustering employing  instantaneous position as observation and trajectory as a mixture of topics to extract semantic regions within a scene.
    
     The work presented in \cite{IncrementalDPMM} performs trajectory clustering using Time-Sensitive Dirichlet Process Mixture Model (tDPMM)~\cite{zhu2005time}. The authors of \cite{xu2015unsupervised} have used adaptive multi-kernel-based estimation together with K-means to produce accurate clusters. The work in~\cite{2} uses DPMM~\cite{Rasmussen} to extract repeating patterns (Motifs) and their occurrence time.
	
A learned model representing  trajectory patterns can be used for different purposes. It can be used for abnormality detection~\cite{16,17,18}, or classification and abnormality detection together~\cite{4,8}. Trajectory retrieval~\cite{IncrementalDPMM,1} is another possible application. The learned models can be used for online classification and abnormality detection even with partially observed tracks. The problem has been addressed in~\cite{4,8,9}. Recently,  Deep Neural Network (DNN)-based research has gained significant momentum. UAV-based traffic flow monitoring presented in~\cite{2017_Zhang_Traffic_Monitoring_DNN} uses deep learning for object detection and recognition. Deep learning has been used for traffic flow prediction using the data from loop detectors  \cite{Deep_trafficflow_2015_Loop}. Multi-sensor user data have been used  for traffic flow forecasting \cite{zhao2017lstm}. In~\cite{2016_Speed_deep}, traffic speed has been predicted with the help of DNN using Global Position System (GPS) data of vehicles.


\subsection{Motivation and Research Contributions}
\label{chap4_sec:Motivation} 

\begin{figure}[h]
  \centering
  \subfigure[Tracks from QML dataset]{\includegraphics[scale=0.33]{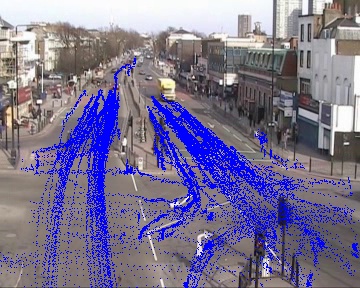}}
  \subfigure[Start/End point distribution]{\includegraphics[scale=0.44]{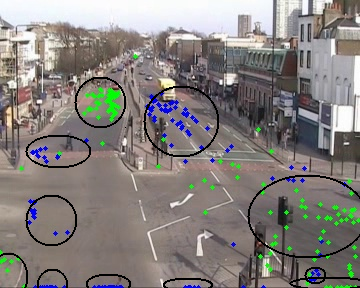}}
  \caption{(a) Shows the tracks of objects in a typical traffic scene taken from QML dataset. (b) Represents the start (blue) and end points (green) of the tracks.}
  \label{chap4_Fig:TrackDist}
\end{figure}

\begin{figure}[!h]
  \centering
      \subfigure[DPMM]{\includegraphics[scale=0.45]{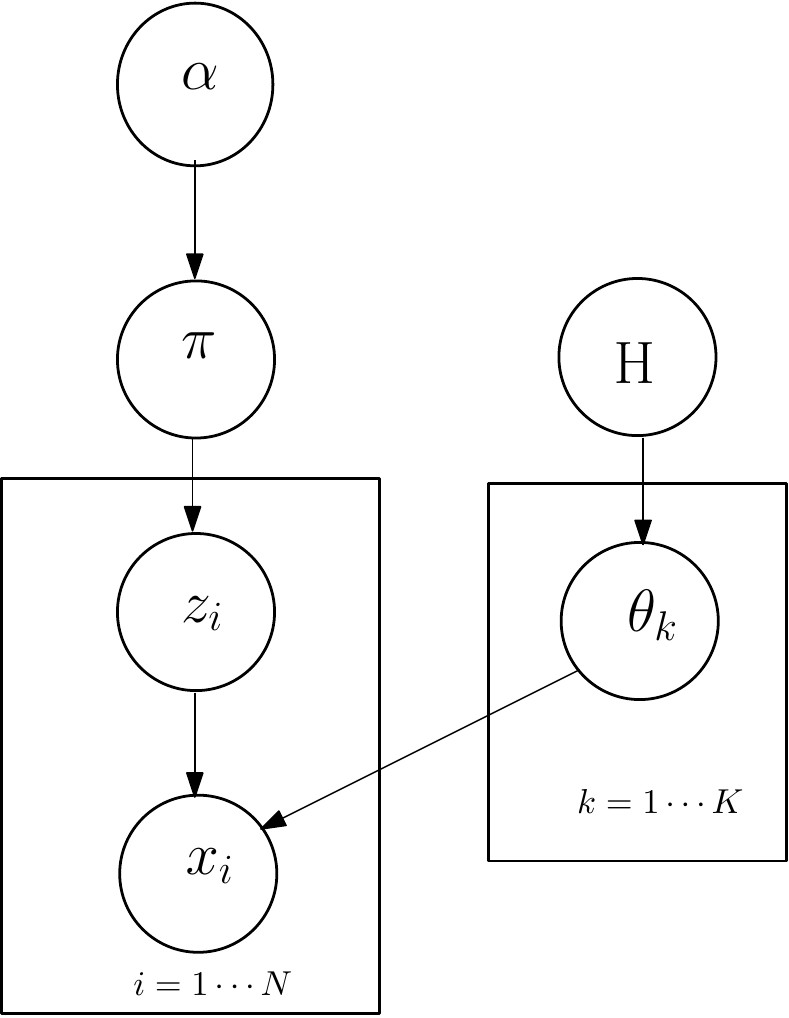}}
      \subfigure[Modified DPMM]{\includegraphics[scale=0.65]{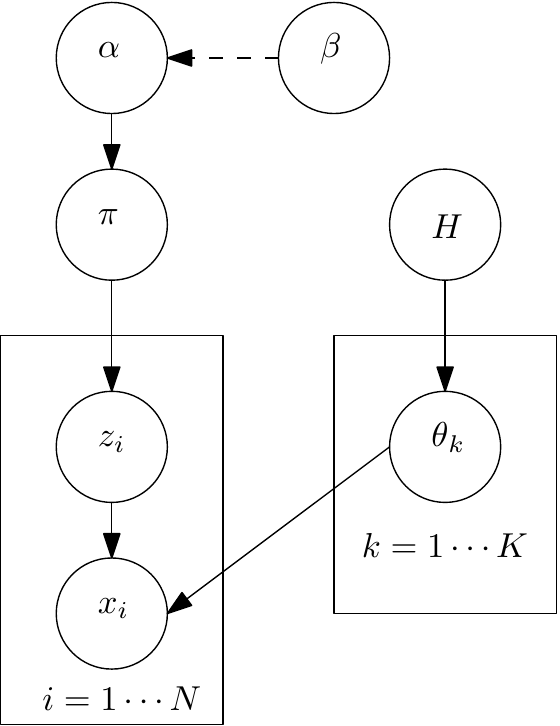}}
    \caption{(a) A conventional Dirichlet Process Mixture Model (DPMM). (b) Modified DPMM using  parameter $\beta$.}
  \label{chap4_Fig:DPMM}  
\end{figure}

 The distributions of start and end points of tracks depicted in Fig.~\ref{chap4_Fig:TrackDist} reveal that the tracks can be grouped using their spatial closeness.  If the likelihood function can model the points in close proximity as a cluster, a nonparametric grouping of tracks can be done. Conventional DPMM, as expressed in ~(\ref{chap4_equation:DPM1}~-~\ref{chap4_equation:DPM4}) can be a good model for nonparametric modeling of data.  
  
\begin{equation}
z_i|\pi  \sim  \mbox{Discrete}(\pi)
\label{chap4_equation:DPM1}
\end{equation} 
\begin{equation}
x_i|z_i, \theta_k  \sim  F(\theta_{z_i})
\label{chap4_equation:DPM2}
\end{equation} 
\begin{equation}
\pi = (\pi_1, \cdots,\pi_K)|\alpha  \sim  \mbox{Dirichlet}(\alpha / K, \cdots, \alpha / K)
\label{chap4_equation:DPM3}
\end{equation} 
\begin{equation}
\theta_k|H  \sim  H
\label{chap4_equation:DPM4}
\end{equation} 
Here, $x_i$ is a random variable representing the data and $z_i$ corresponds to the latent variable representing cluster labels, where $i = 1 \cdots N$ with $N$ being the number of data points. $z_i$ takes one of the values from $k = 1 \cdots K$, where $K$ is the number of clusters. $\pi= (\pi_1, \cdots,\pi_K)$, referred to as mixing proportion, is a vector of length $K$ representing the probabilities of $z_i$ to be $k$. $\theta_k$ is the parameter of  cluster $k$ and $F(\theta_{z_i})$ denotes the distribution defined by $\theta_{z_i}$. $\alpha$ denotes the concentration parameter and its value decides the number of clusters formed. Initially,  $z_i$ is picked from a Discrete distribution given in~(\ref{chap4_equation:DPM1}). Next, the data points are generated from a distribution parameterized by $\theta_{z_i}$, as given in~(\ref{chap4_equation:DPM2}). Here, $\pi$ is derived from a Dirichlet distribution as given in~(\ref{chap4_equation:DPM3}) and $\theta_k$ is derived from the distribution $H$ of priors as represented in~(\ref{chap4_equation:DPM4}). The model~\cite{GraphicModel} is graphically presented in Fig.~\ref{chap4_Fig:DPMM}(a). In ~\cite{TUIC_ITS_2018}, a method has been proposed to express the concentration parameter ($\alpha$) in terms of distance using the deterministic relation $\alpha = e^{-\beta}$, where $\beta$ is referred to as the concentration radius. 

 However, the modified DPMM as shown in Fig.~\ref{chap4_Fig:DPMM}(b) cannot be directly used for representing trajectory data as timing information is not considered in this model. In the proposed scheme, temporal correlations of data points are considered rather than representing the trajectories as a collection of unrelated data. Moreover, as the concentration parameter is represented in terms of a distance function, it can provide an intuition on how to choose $\alpha$ for learning the model parameters representing the dynamics of a traffic scene. It has been observed that a single iteration of Gibbs sampling~\cite{25} is sufficient to associate tracks to clusters by learning the most frequently used paths in a given scene. In the proposed DEM, the learned parameters change temporally, thus allowing us to detect unusual events without revisiting the whole data points, unlike other statistical methods~\cite{KMeansJin2010, comaniciu2002mean, ester1996density}. 
 
  Categorization of events as normal can be highly contextual and it has some relation with time. For example, during busy traffic, the speed of a vehicle can be reasonably slow. If a person drives significantly faster as compared to other vehicles, it can be termed abnormal. However, in sparse traffic, similar speed can be termed as normal. This is referred to as the scene dynamics and it has been diligently addressed in the proposed model. Another important characteristic is the temporal aspect, i.e., objects moving close to each other around the same time have similar temporal correlations. This can be understood from the following example. Vehicles at the front of the queue start moving followed by the vehicles behind when the signal turns green. Moreover, there is a temporal relation between entry and exit events of the objects. These aspects are considered in the proposed clustering, unlike other nonparametric, unsupervised methods such as DBSCAN~\cite{ester1996density} or mean shift~\cite{comaniciu2002mean}.

The main objective of this paper is to understand the spatio-temporal characteristics of the usual patterns from the moving object trajectories. This can be a building block for traffic behavior analysis. While accomplishing the above objective, the following contributions have been made in this paper:

\begin{enumerate}[(i)]
\item A DPMM guided model that is referred to as Temporally Incremental Gravity Model (TIGM) and an inference scheme to create spatio-temporal clusters using conditional dependencies among the trajectories have been proposed.
\item As typical DPMM clustering depends on the concentration parameter ($\alpha$) that is difficult to estimate, a method to represent $\alpha$ using a distance-based measure has been proposed.
\item A model namely, Dynamically Evolving Model (DEM) that reflects the variations in the scene parameters has been proposed by temporally extending the TIGM.
\item Moreover, how the proposed DEM can be used to design traffic analysis frameworks has been presented. 
\end{enumerate}

Rest of the paper is organized as follows. In Section~\ref{chap4_Methodology}, rationale for building the proposed methodology is discussed. In Section~\ref{chap4_sec:Results}, we discuss experimental setup, datasets and parameters. In addition, we analyze and summarize the results. In Section~\ref{chap4_sec:Discussion}, we discuss other relevant methods compared to the proposed method. We have also discussed a few limitations of the proposed methods. Section~\ref{chap4_sec:conclusion} concludes our work with peek into the future directions. 
\section{Proposed Methodology}
\label{chap4_Methodology}
At first, we discuss the terminologies used in the paper. The terms \textit{observation} and \textit{data} are used interchangeably. Here, they represent the tracks or trajectories. Similarly, we refer \textit{cluster} or \textit{topic} to denote distribution of data and is represented by a label. They indicate most frequently used paths. A model is used to represent a real-world phenomenon. We use graphical model~\cite{GraphicModel} for representing the mixture models. A graphical model represents the generative model of the data. A parametric model has a fixed number of parameters, while the number of parameters grows with the amount of training data used to construct the nonparametric models. We follow an unsupervised method for learning a scene and our model is nonparametric. This characteristic is essential to understand the scene dynamics, which is defined later.
 \subsection{Temporally Incremental Gravity Model (TIGM)}
\label{chap4_sec:Problem_formulation}
  The following assumptions have been applied to the proposed traffic analysis framework: 
 \begin{itemize}
\item[(i)] The start or end points of the objects in motion constitute a statistical distribution of data in 2D.
\item[(ii)] If tracks are clustered based on the proximity, frequently used paths can be found.  
\end{itemize}
To start with, the inference process used in TIGM is discussed. Here, $x_i$ represents $i^{th}$ trajectory and $z_i$ the corresponding cluster label. The goal is to find $z_i$ for all tracks.  $z_i$ corresponds to the cluster label $k \in \{1 \cdots K\}$. $x_i$ is  associated with a cluster of unknown distribution parameterized by $\theta_k = \{\mu_k, \Sigma_k\}$. Here, $\mu_k$ and $\Sigma_k$ represent the mean and covariance of the distribution, respectively. 

The inference method can be expressed using~(\ref{chap4_equation:INF1}-\ref{chap4_equation:INF2})  \cite{25}, where $()_{-i}$ corresponds to the parameters excluding the $i^{th}$ data. Here, (\ref{chap4_equation:INF1}) represents the probability of an observation forming a new cluster and (\ref{chap4_equation:INF2}) represents the probability of an observation belonging to one of the K existing clusters. $F(x_i|\theta_k)$ corresponds to the likelihood of $x_i$ in $\theta_k$.  $K = 0$, when no observation is present. $n_{-i}$ denotes the number of observations for which labels are already assigned, excluding $x_i$. $\alpha$ represents the concentration parameter of the Dirichlet distribution. These equations define the posterior probability in terms of likelihood and the prior. In other words, the probability of an observation belonging to a cluster $k$ (given other parameters) is proportional to the probability of the observation being generated from the cluster $k$ multiplied by the proportion of the observations present in $k$ as given below.
\begin{equation}
p(z_i = K+1 | z_{-i},x_i, \theta_{k_{-i}}, \alpha) \propto  F(x_i| \theta_{K+1}) \times\frac{\alpha}{n_{-i} + \alpha}
\label{chap4_equation:INF1}
\end{equation} 
\begin{equation}
p(z_i = k | z_{-i},x_i, \theta_{k_{-i}}, \alpha) \propto  F(x_i| \theta_k) \times \frac{n_{k_{-i}}}{n_{-i} + \alpha}
\label{chap4_equation:INF2}
\end{equation}   
The likelihood function needs to be selected in such a way that the probability of observation to be associated with a cluster increases when the observation is nearer to the cluster mean. The association probability gradually reduces as the observation moves away from the cluster mean. The exponential decay function $e^{-x}$ follows the above characteristics~\cite{TUIC_ITS_2018}. Here, $x$ represents the distance of $x_i$ from $\mu_k$. The inference equation can be written as (\ref{chap4_equation:INF5_INT}). $k = K+1$ denotes the probability of the observation forming a new cluster and the later part signifies the probability of $x_i$ belonging to one of the $K$ existing clusters.
\begin{align} \label{chap4_equation:INF5_INT}
p(z_i = k | z_{-i},x_i,\theta_{k_{-i}},\alpha)
&\propto \begin{cases}
 e^{-x} \times \frac{\alpha}{n_{-i} + \alpha}, \text{if $k = K + 1$;}\\
 e^{-x} \times \frac{n_{k_{-i}}}{n_{-i} + \alpha},\text{otherwise.}
\end{cases}
\end{align} 
 When an observation forms a new cluster, the likelihood function gets reduced to $e^{-x} = 1$ since the distance of an observation to itself is $0$. The inference equation can be further simplified to~(\ref{chap4_equation:INF5}) by removing the proportionality symbol and the common denominator ${n_{-i} + \alpha}$, where $b$ is a normalization constant and $\alpha$ = $e^{-\beta}$. $\beta$ is referred to as the concentration radius. This inference process is derived without assuming $K \to \infty$, unlike \cite{Rasmussen,25,ishwaran2002exact}.
\begin{align} \label{chap4_equation:INF5}
p(z_i = k | z_{-i},x_i,\theta_{k_{-i}},\beta)
&= \begin{cases}
b \times e^{-\beta}, \text{ if $k = K + 1$;}\\
b \times e^{-x} \times n_{k_{-i}},\text{otherwise.}
\end{cases}
\end{align} 
 The above equation is the key to find the value of the concentration parameter ($\alpha$) as $\beta$ can be expressed in terms of distance from the center to a point at the periphery of the distribution. If an observation creates a new cluster, it has to be at an infinitesimal distance ($\delta x$) higher than that of the distribution's periphery from the mean ($x = max$). This implies, $e^{-\beta} = e^{-(max + \delta x)} \times n_{k_{-i}}$. Therefore, the value of $\beta$ can be estimated using~(\ref{chap4_eq:beta_initial}).
\begin{equation}
\label{chap4_eq:beta_initial}
\beta =  (max + \delta x) - \ln(n_{k_{-i}}).
\end{equation}
It may be observed that $\delta x$ can be ignored since $\delta x \textless \textless max$. Also, $\ln(n_{k_{-i}}) \approx \ln(n_{k})$, yields $\beta = max - \ln(n_k)$.  Once the initial clustering is done using an approximate $max$ corresponding to the radius of the data distribution, the actual $max$ can be obtained with some fine-tuning of $n_k$. Thus, a better clustering result can be obtained.  
 
\begin{figure}[!t]
  \centering
 \subfigure[TIGM]{\includegraphics[scale=0.75]{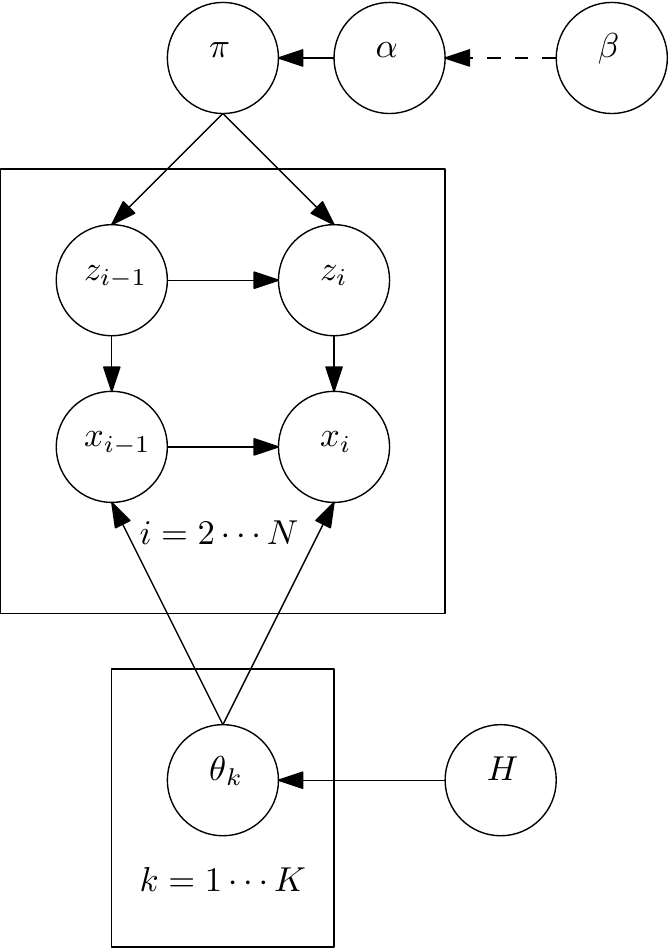}}
 \subfigure[TIGM Illustration]{\includegraphics[scale=0.42]{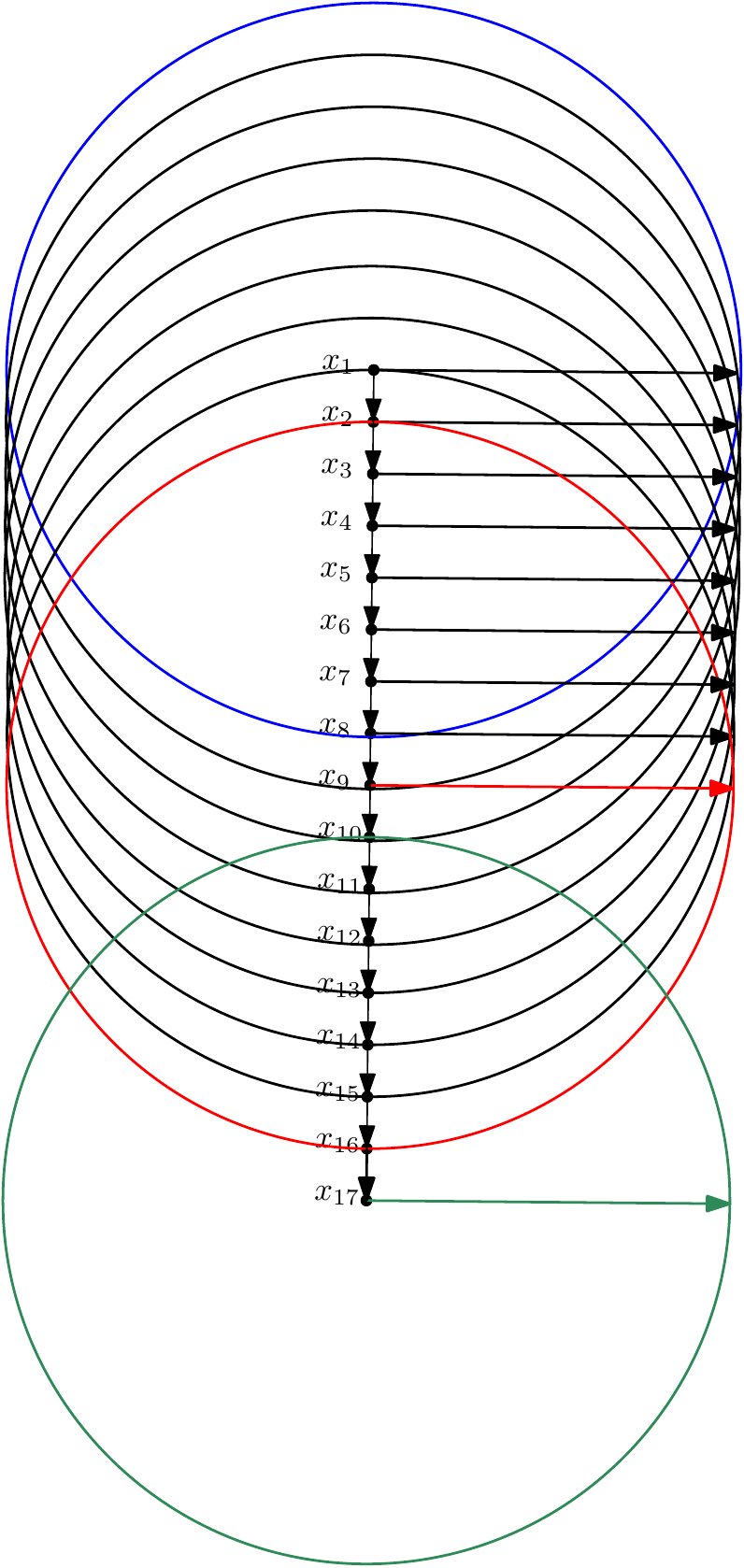}}
    \caption{(a) Temporally Incremental Gravity Model (TIGM). Here, the dashed line represents the deterministic relation, $\alpha$ = $e^{-\beta}$. (b) A depiction of TIGM using data points in a straight line. If the sampling starts from top to bottom ($x_1$ to $x_{17}$), though the concentration radius does not fit all the points, they will be grouped into one cluster due to the strength parameter ($n_k$).  If sampling starts randomly, it may form clusters with centers at $x_1$, $x_9$, and $x_{17}$. This aspect is represented in TIGM through conditional dependencies from $x_{i-1}$ to $x_i$.}
  \label{chap4_Fig:TIGM}    
\end{figure}

 The distance ($x$) can be chosen as Euclidean distance, Mahalanobis distance, dynamic time warping (DTW)~\cite{berndt1994using}, etc. Here, Euclidean distance has been used though Mahalanobis can also approximate elliptical distributions. This has been done intentionally to limit the distribution variance in $XY$ plane. However, depending on the features and applications, a suitable distance measure can be used. To simplify the model, a fixed $\beta$ has been used, though different distributions may have varying spreads. Hence the posterior probabilities are learned, not the likelihood function. The model is represented graphically in Fig.~\ref{chap4_Fig:TIGM}(a). 

  The model is referred to as Temporally Incremental Gravity Model (TIGM) for the following reasons: (i) The clusters are able to attract more observations from the neighborhood. This is not only due to the spatial proximity but also for incremental temporal order. Thus, clustering does not strictly become bounded by the  distance. The gravitational strength (a large group of closely appearing observations attracts more nearby observations) increases when more observations are added to the cluster. In order to illustrate this idea, consider a list of points equally apart over a straight line segment as shown in Fig.~\ref{chap4_Fig:TIGM}(b). The mean of the data points will shift down if the data points are sampled incrementally from top to bottom. If a decision is made to cluster the points strictly based on a radius of half the length of the line segment, it may form more than one clusters depending on the order of samples being drawn. If done incrementally, it will form exactly one cluster. Moreover, even if a radius less than half the line segment is considered, the model can form a single cluster as mixing proportion factor attracts more observations to the cluster. As explained earlier, when a signal turns green at a traffic junction, vehicles at the beginning usually move earlier. Unless front vehicles move, vehicles behind cannot move. This is the rationale behind modeling the conditional dependencies in TIGM. It means clustering based on the temporal order of arrival can yield the best results as observation dependencies are included in the computation. In typical statistical approaches, where all observations are considered for clustering, multiple iterations of sampling may be required to achieve similar clustering results. (ii) Temporal information is maintained as the inference is done in an incremental fashion. This model is different from DPMM as both temporal correlation between the observations and corresponding latent variables using the conditional dependence are represented within the model. The generative process can be represented using~(\ref{chap4_equation:TIGM1}-\ref{chap4_equation:TIGM4}).
\begin{equation}
z_i|\pi,z_{i-1}  \sim  \mbox{Discrete}(\pi)
\label{chap4_equation:TIGM1}
\end{equation} 
\begin{equation}
x_i|z_i, x_{i-1}, \theta_k  \sim  F(\theta_{z_i})
\label{chap4_equation:TIGM2}
\end{equation} 
\begin{equation}
\pi |e^{-\beta}  \sim  \mbox{Dirichlet}(e^{-\beta} / K, \cdots, e^{-\beta} / K)
\label{chap4_equation:TIGM3}
\end{equation} 
\begin{equation}
\theta_k|H  \sim  H
\label{chap4_equation:TIGM4}
\end{equation} 
\subsection{Dynamically Evolving Model (DEM)}
\label{chap4_sec:Problem_formulation_2}
TIGM cannot estimate traffic trend of a scene across different time segment as scene dynamics is not included in the model. The model is extended to reflect the temporal characteristics of a scene. Before proceeding further, scene dynamics needs to be quantified. It is denoted by $\Phi^t$ and  defined as the parameter vector of length $K^t$ such that $\Phi^t = <\Phi_1^t, \cdots, \Phi_{K^t}^t>$. Here, $\Phi_k^t$ represents cluster dynamics in the $t^{th}$  temporal time segment of duration $\Delta t$ for the cluster $k^t$. It is defined as $\Phi_k^t = <\mu_k^t, \Sigma_k^t, n_k^t>$, where $\mu_k^t$ denotes the mean of the cluster $k^t$, $\Sigma_k^t$ denotes its covariance and $n_k^t$ denotes the number of observations in cluster $k^t$. The model becomes complete with the following additional assumptions:
\begin{itemize}
\item[(i)]When data is captured using a static camera, the cluster statistics do not change significantly in short duration.
\item[(ii)]For longer intervals, the traffic characteristics can be significantly different and clustering over longer windows can be meaningful for real-time monitoring systems.
\item[(iii)]If time segment is fixed and clustering is done with prior information about the learned cluster, semi-supervised learning can be used for clustering and classification of the trajectories in subsequent temporal segments.
\end{itemize}     
Based on the above assumptions, new trajectories can be added and old trajectories that are not in the time segment, can be removed from the cluster during learning. If Gibbs sampling is performed using  $\theta_k^{t-1}$ as a prior for the ${t}^{th}$ frame, cluster labels can be maintained between consecutive frames. The inference can be realized using ~(\ref{chap4_equation:INFT}) with exactly one iteration of the Gibbs sampling. The rationale behind using a single iteration is to ensure that the temporal aspects are not lost during clustering. If re-sampling is done at known intervals, stabilization of scene dynamics can be achieved. Here, $z_{-i}^*$ is different from the $z_{-i}$ discussed earlier. It represents the set of all cluster assignments except for $x_i^{t}$ so as to exclude the observations and corresponding cluster assignments before ${t-1}$. $\theta_{k_{-i}}^{*}$ is the parameter representing the distribution corresponding to cluster $k^t$  by excluding observation $x_i^{t}$. Here,  $\mu_{-i}^{*}$ is the mean of $\theta_{k_{-i}}^*$ distribution. $n_{k_{-i}}^*$ denotes the number of observations in $\theta_{k_{-i}}^*$ and $b$ is the normalization constant. Thus, 
\begin{align} \label{chap4_equation:INFT}
p(z_i^{t} = k | z_{-i}^*,x_{i}^t,\theta_{k_{-i}}^*,\beta)
&= \begin{cases}
b \times e^{-\beta},        \text{ if $k = K + 1$;}\\
b \times e^{-x} \times n_{k_{-i}^*},  \text{ otherwise.}
\end{cases}
\end{align} 
\begin{figure}[!tbp]
  \centering
      \includegraphics[scale=0.7]{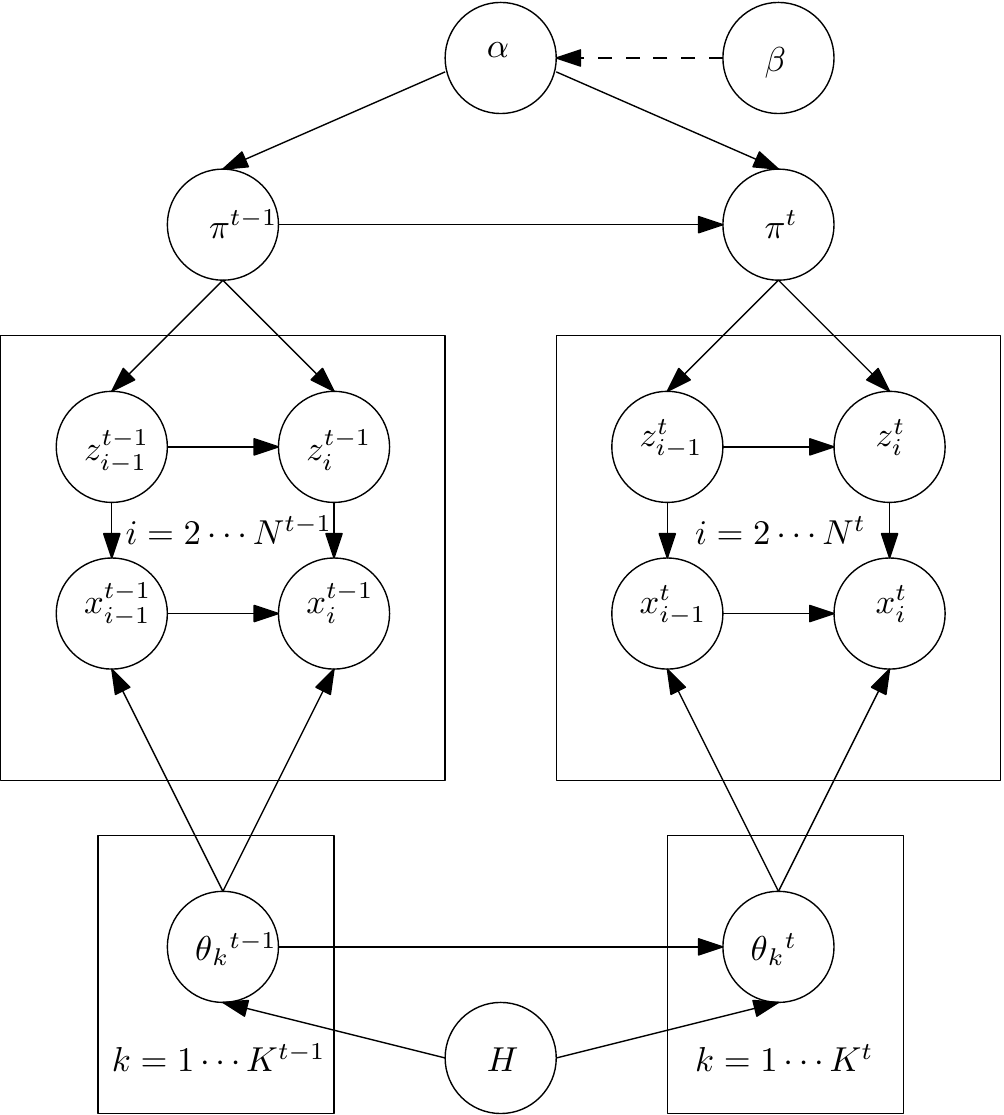}
    \caption{Dynamically Evolving Model (DEM) depicting how clustering can vary temporally. The dashed line represents the deterministic relation, $\alpha$ = $e^{-\beta}$.}
  \label{chap4_Fig:DEM}  
\end{figure}
The proposed model is shown in Fig.~\ref{chap4_Fig:DEM}. It can be represented  using~(\ref{chap4_equation:ITCM1}-\ref{chap4_equation:ITCM4}). Here, $x_i^t$   ($i = 1 \cdots N_t$) corresponds to the observation at $t$ excluding the observations before $t-1$ and $z_i^t$ ($i = 1 \cdots N_t$) corresponds to the latent variable representing  cluster labels, taking one of the values from $k = 1 \cdots K^t$. $N_t$ is the number of observations and $K^t$ is the number of clusters at $t$. $\pi^t$ is a vector of length $K^t$. $\pi_k^t = (\pi_1^t, \cdots,\pi_K^t)$ represents the mixing proportion of observations among the clusters. $\theta_k^t$ is the parameter of the $k^{th}$ cluster and $F(\theta_{z_i^t})$ denotes the distribution defined by $\theta_k^t$. Initially, $z_i^t$ is picked from a Discrete distribution given in~(\ref{chap4_equation:ITCM1}). The data is then generated from a distribution parameterized by $\theta_{z_i}^t$ as given in~(\ref{chap4_equation:ITCM2}), where $\pi^t$ is derived from a Dirichlet distribution as given in (\ref{chap4_equation:ITCM3}). $\theta_k^t$ is derived from another distribution $H$ of prior as represented in~(\ref{chap4_equation:ITCM4}). The model is different from the original TIGM shown in Fig.~\ref{chap4_Fig:TIGM}. Here, $\theta_k^t \to \theta_k^{t-1}$ and $\pi^t \to \pi^{t-1}$. This is meaningful as the distributions do not change significantly within short intervals. For example, vehicle density usually does not vary significantly within short interval (in a span of 4-5 minutes) during peak hours. Thus, there is a conditional dependence of both $\theta_k$ and $\pi$ in between successive time slots.
\begin{equation}
z_i^t|\pi^t,z_{i-1}^{t}  \sim  \mbox{Discrete}(\pi^t)
\label{chap4_equation:ITCM1}
\end{equation} 
\begin{equation}
x_i^t|z_i^t,x_{i-1}^{t}, \theta_{k}^t  \sim  F(\theta_{z_i^t})
\label{chap4_equation:ITCM2}
\end{equation} 
\begin{equation}
\pi^t |e^{-\beta},\pi^{t-1}  \sim  \mbox{Dirichlet}(e^{-\beta} / K^t, \cdots, e^{-\beta} / K^t)
\label{chap4_equation:ITCM3}
\end{equation} 
\begin{equation}
\theta_k^t|H,\theta_k^{t-1}  \sim  H
\label{chap4_equation:ITCM4}
\end{equation}
\begin{algorithm}[!h]
\caption{Dynamically Evolving Model (DEM)}
\textbf{Input:} Input video, $\beta$, Duration of time segment ($\Delta t$) \\
\textbf{Output:} Motifs
\begin{algorithmic}[1]
\State $K = 0$; $time = 0$; $prevtime = 0$;
\State Start a cyclic timer of duration $\Delta t$ for updating $time$;
\State frame = Next Frame;
\For {!Empty (frame)}
\State Track the objects from previous frame;
\State Create $x_i$ for completed tracks;
\For{each newly created $x_i$ }
\State Find $z_i$ using~(\ref{chap4_equation:INFT});
\EndFor
\If {($time >= prevtime + 1$)}
\State un-assign $x_i$ arrived before $prevtime$;
\State Resample and find $z_i$ using~(\ref{chap4_equation:INFT});
\State Display the frame with cluster labels;
\State $prevtime = time$;
\EndIf
\State frame = Next Frame;
\EndFor
\end{algorithmic}
\label{chap4_algorithm:1}
\end{algorithm} 
The inference method for cluster assignment uses Gibbs sampling~\cite{25}. The process is described in Algorithm~\ref{chap4_algorithm:1}. Here, $x_i$ can be $<x_{start},y_{start},x_{end},y_{end},t_{dur}>$, where $<x_{start},y_{start}>$  and $<x_{end},y_{end}>$ represent start and end positions and $t_{dur}$ represents the duration of the track. The rationale for using only these features (start and end points and trajectory duration) is, a majority of the vehicles follow typical patterns while moving on roads. Therefore, those following similar patterns will be closely distributed in a multi-dimensional space. Also, $x_i$ can be $<x_{start},y_{start}>$, $<x_{end},y_{end}>$, and a combination of both. Then, it can be used to find the tracks originating from and/or ending at a particular region.
  
\section{Experimental Results}
\label{chap4_sec:Results}
\subsection{Experimental Setup and Datasets}
\label{chap4_sec:experimental_setup}
 OpenCV has been used to implement the proposed framework. Experiments have been carried out on four publicly available traffic video datasets, namely UCF~\cite{UCF_DATASET}, QML~\cite{QMUL_DATASET}, MIT~\cite{wang2011trajectory}, and Grand Central Station, New York (GCS)~\cite{zhou2012understanding}. UCF is a traffic junction video of one-minute duration captured from the top of a building covering the vehicle movements across different roadways of a complex junction.  QML is a video dataset of approximately 50 minutes duration of a busy traffic junction showing vehicles and pedestrian movements with a few anomalies. The trajectories are extracted using context tracker~\cite{dinh2011context}. For TIGM experiments on QML, the trajectories of first 10 minutes duration have been used. MIT dataset contains trajectories of moving vehicles inside a parking area captured from the top of a nearby building. GCS is a trajectory dataset containing the tracks of human movements inside an underground metro station, namely Grand Central Station, New York. TIGM experiments on the datasets UCF, QML, MIT,  and GCS are carried out using 76, 146, 449, and 3500 trajectories, respectively. DEM experiments on QML dataset have been conducted by extracting around 16K trajectories using the method present in~\cite{zhou2013measuring}.

\subsection{TIGM Parameter ($\beta$) Variation}
\label{chap4_sec:BETA_Variation}
Deciding a  suitable $\beta$ for clustering is intuitive for a particular scene. Our analysis reveals that for start/end points-based clustering, an initial $max$ value can be an approximate radius in the X-Y plane corresponding to the start/end point distribution. Choosing a suitable value of $max$ has significant influence in clustering, as $\ln(n_k)$ is smaller as  compared to $max$. Refinement has been done using the term $\ln(n_k)$ depending on the number of trajectories in the clusters. For higher dimensions, initial $max$ value is set as two times (with additional two dimensions) or three times (with additional three dimensions) the refined $\beta$ value that has been obtained using the previous step. Further refining can be done similar to the method applied for lower dimensions.

\begin{figure}[!tbp]
  \centering
      \includegraphics[scale = 0.43]{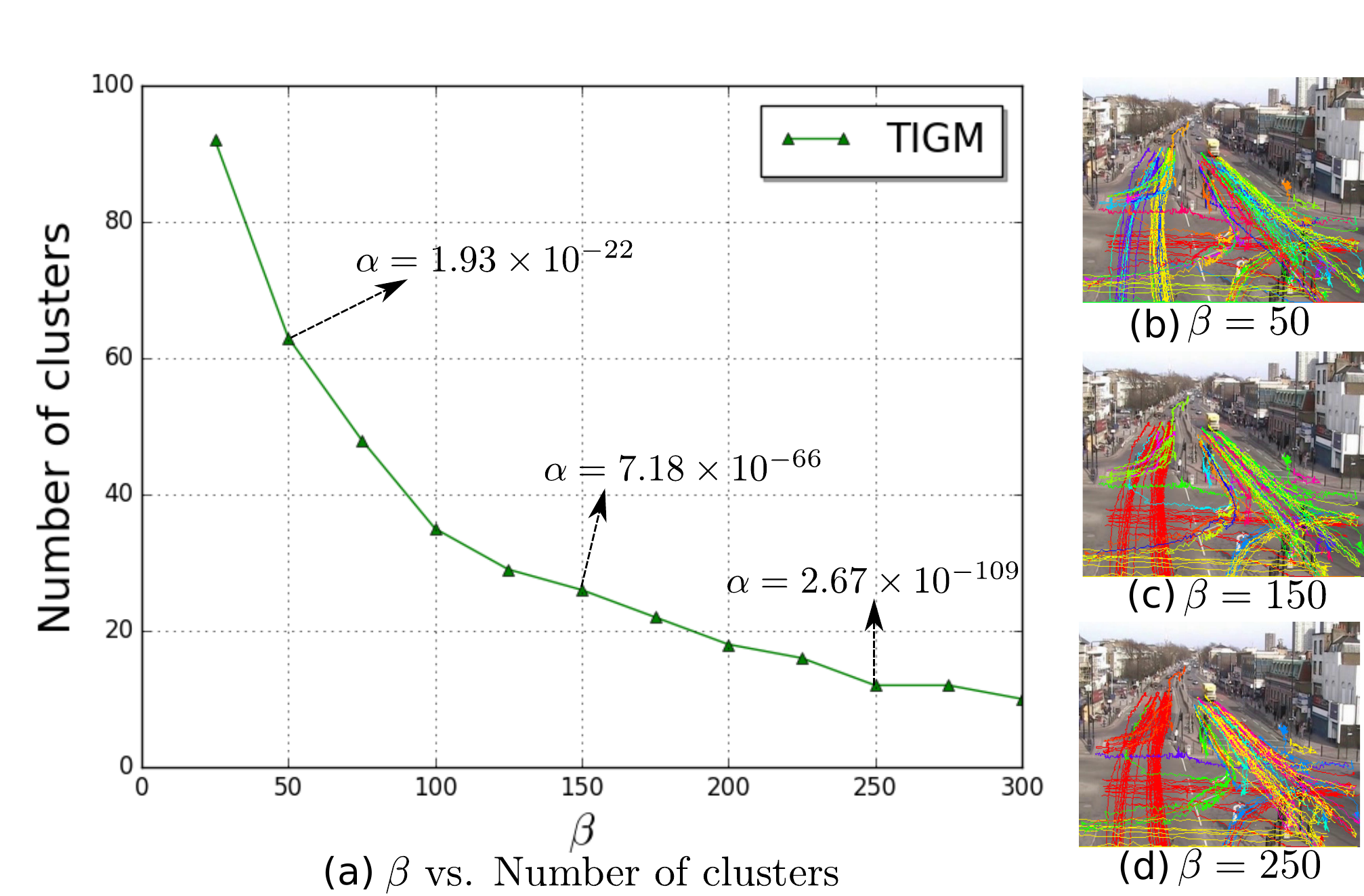}
    \caption{Depiction of how $\beta$  influences the number of clusters formed. (a) Plot of $\beta$ vs. number of clusters. It also shows approximate $\alpha$ values for some $\beta$. (b - d) The clustered trajectories for three different $\beta$ values.}
  \label{chap4_Fig:BETA_VS_CLUSTERS}  
\end{figure}

As the concentration parameter ($\alpha$) is represented in terms of concentration radius ($\beta$), it is important to understand how $\beta$ variation influences the clustering results. The results shown in Fig.~\ref{chap4_Fig:BETA_VS_CLUSTERS} suggest that there is an inverse relation between $\beta$ and the number of clusters formed. However, the relation is more intuitive as compared to that of the relation between $\alpha$ and the number of clusters. The $\alpha$ is difficult to estimate as its value is typically less than $10^{-20}$.

\subsection{TIGM Experiments and Analysis}
\label{chap4_sec:Convergence}
\begin{figure}[!h]
  \centering
 \subfigure[TIGM]{\includegraphics[width=0.155\textwidth]{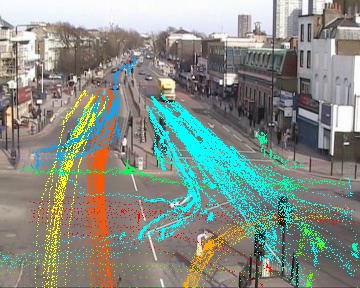}}
 \subfigure[TIGM converged]{\includegraphics[width=0.155\textwidth]{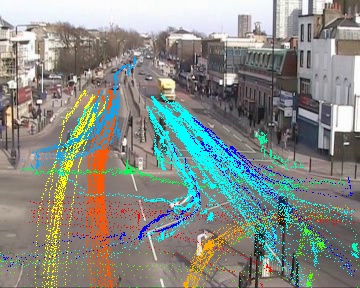}}
 \subfigure[cluster 9]{\includegraphics[width=0.155\textwidth]{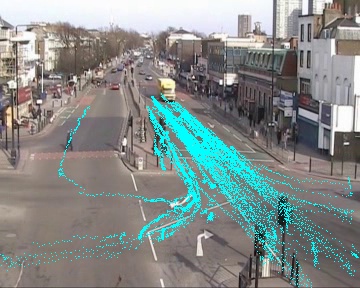}}      
  \subfigure[Converged cluster 12]{\includegraphics[width=0.155\textwidth]{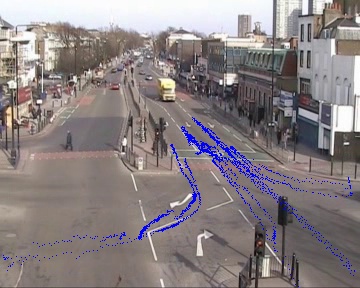}}
    \subfigure[Converged cluster 2]{\includegraphics[width=0.155\textwidth]{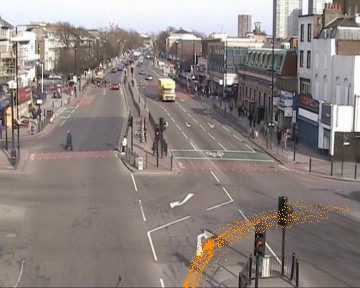}}  
    \subfigure[Track 38]{\includegraphics[width=0.155\textwidth]{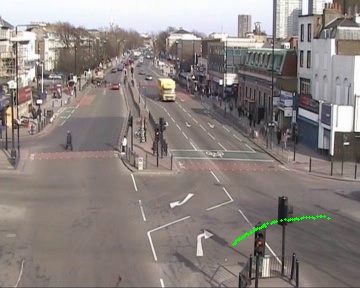}}        
  \caption{Comparisons of trajectory clustering using start points based on convergence. (a) Clusters formed with single iteration using TIGM. (b) Clusters after convergence of model parameters using TIGM. (c) Cluster 9 using TIGM. (d) Cluster 12 using TIGM with convergence. (e) Cluster 2 using TIGM with convergence. (f) Cluster 6 using TIGM with convergence containing only Track 38.}
  \label{chap4_Fig:ConvergenceIssue}
\end{figure}
\begin{figure}[!h]
  \centering
      \includegraphics[scale = 0.35]{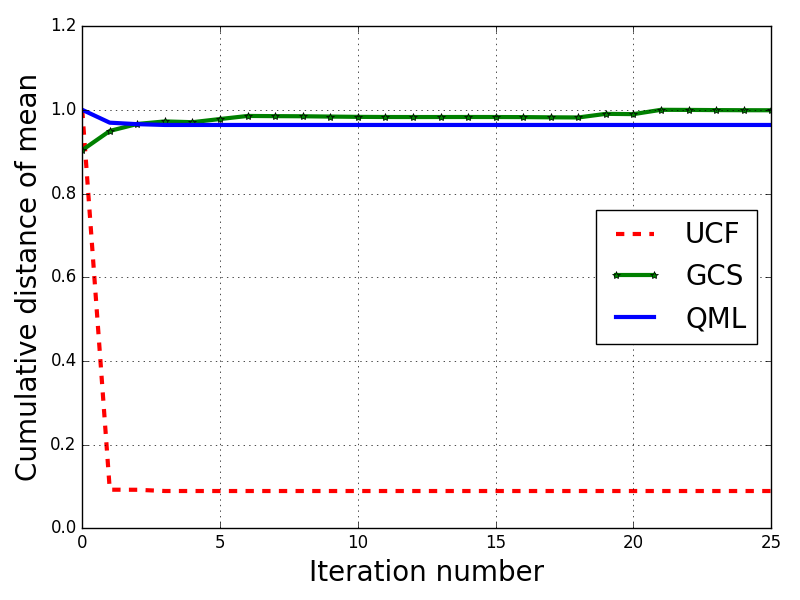}
    \caption{Depiction of the convergence process for three datasets. Y-axis shows normalized cumulative distance of mean from the origin. QML uses start point-based clustering, while in UCF and GCS, all features have been used.}
  \label{chap4_Fig:Convergence}  
\end{figure}
The experiments on three datasets, namely QML, UCF, and GCS reveal, temporal correlation of the trajectories is more important than the convergence of model parameters for continuous monitoring applications. To illustrate this, start point-based clustering has been applied on the QML dataset. The results of TIGM without and with convergence are presented in Figs.~\ref{chap4_Fig:ConvergenceIssue} (a) and (b), respectively. It reveals,  some tracks of cluster $9$ form a new cluster ($12$) with convergence as shown in Fig.~\ref{chap4_Fig:ConvergenceIssue} (d). Similarly, track $38$ becomes part of a new cluster $6$. Though such a grouping can be accepted due to the high spatial distance of track $38$ from cluster $2$, tracks of cluster $12$ are ideally expected to belong to cluster $9$. Such incorrect clustering happens due to the loss of temporal information with convergence. The convergence plots for QML, UCF, and GCS datasets are shown in Fig.~\ref{chap4_Fig:Convergence}. GCS takes more time to converge due to lack of restriction on people movement unlike that of a typical roads, leading to multiple clusters. Though convergence using multiple iterations is relatively faster, clustering results are not meaningful. Hence a single iteration of Gibbs sampling has been adopted. 
\subsubsection{TIGM Experiments and Analysis on Traffic Data}
\begin{figure}[!h]
\centering
 \subfigure[Clustered Tracks]{\includegraphics[width=0.158\textwidth]{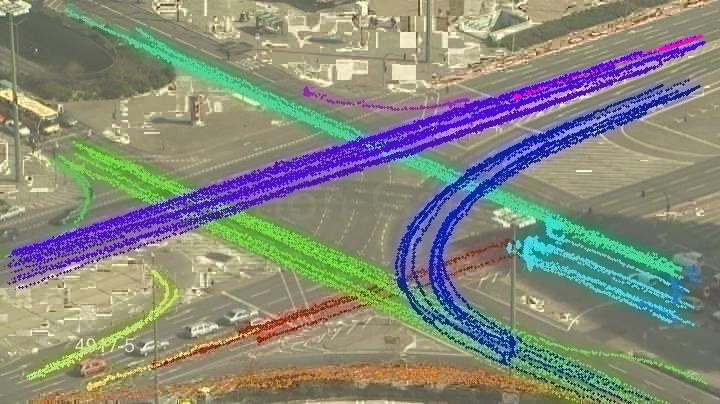}}
 \subfigure[cluster 1]{\includegraphics[width=0.158\textwidth]{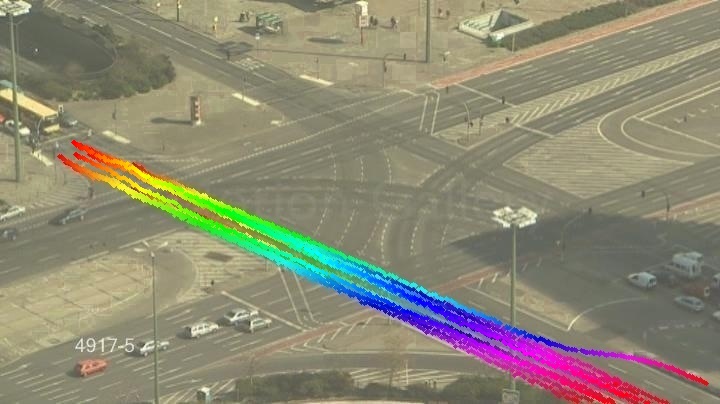}}
 \subfigure[cluster 2]{\includegraphics[width=0.158\textwidth]{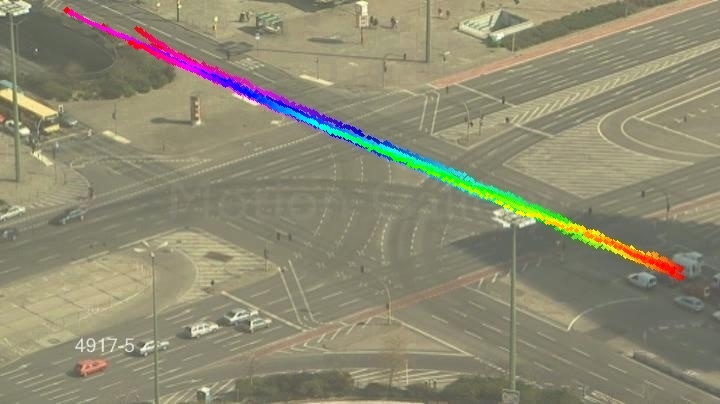}}
 \subfigure[cluster 3]{\includegraphics[width=0.158\textwidth]{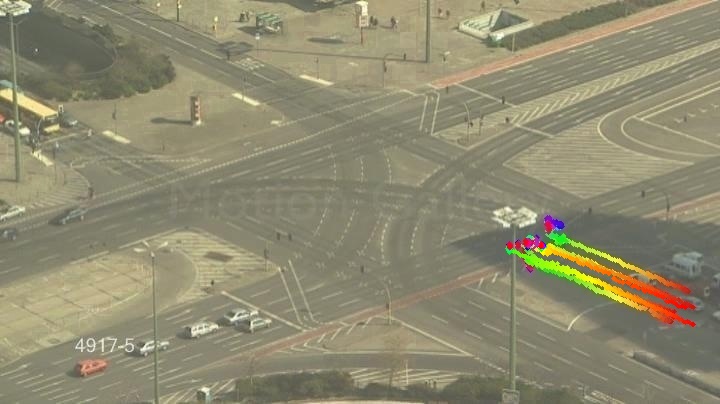}}
 \subfigure[cluster 4]{\includegraphics[width=0.158\textwidth]{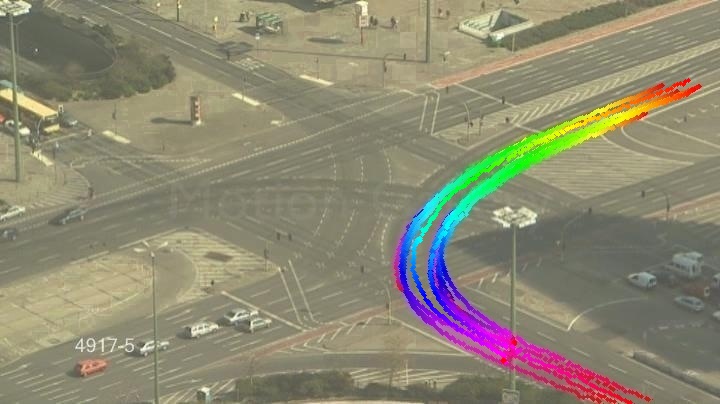}}
 \subfigure[cluster 5]{\includegraphics[width=0.158\textwidth]{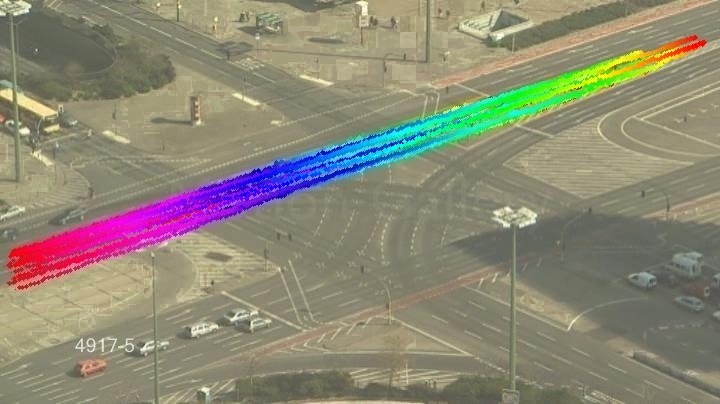}}
 \subfigure[cluster 6]{\includegraphics[width=0.158\textwidth]{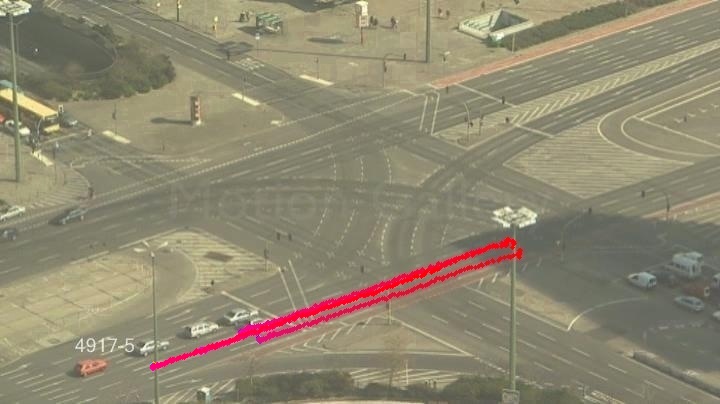}} 
 \subfigure[cluster 7]{\includegraphics[width=0.158\textwidth]{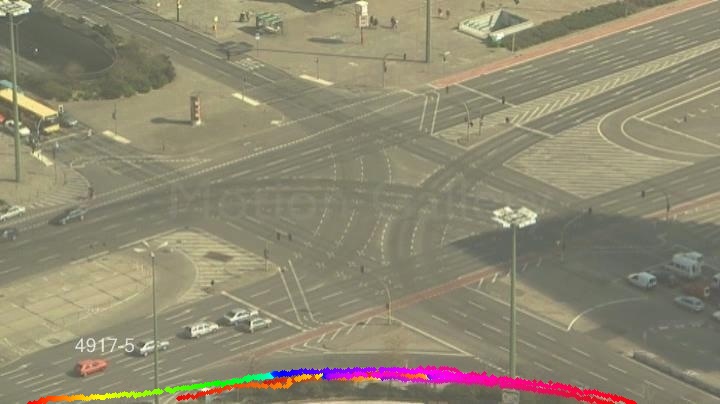}}  
 \subfigure[cluster 8]{\includegraphics[width=0.158\textwidth]{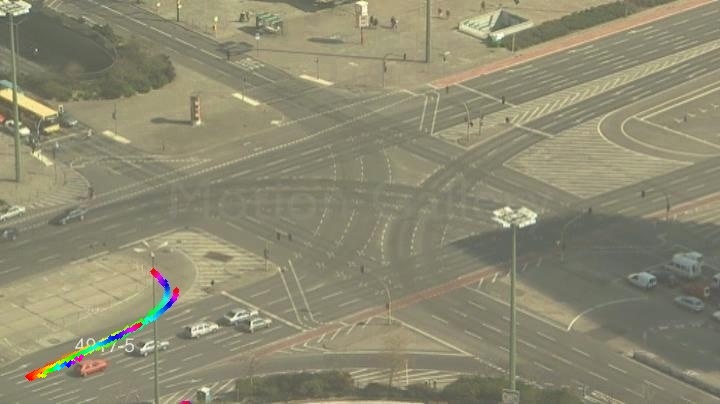}} 
 \subfigure[cluster 9]{\includegraphics[width=0.158\textwidth]{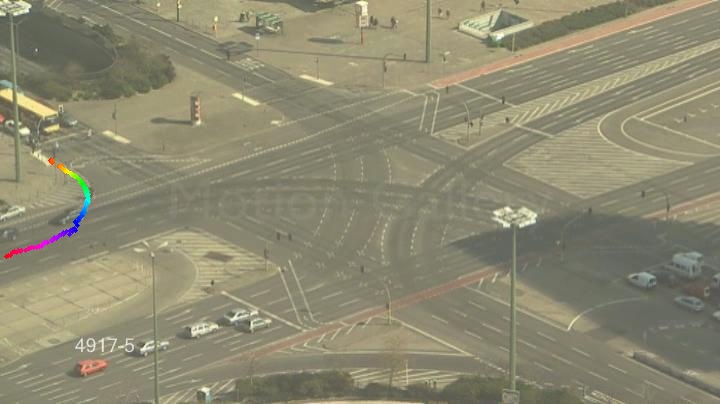}}  
 \subfigure[cluster 10]{\includegraphics[width=0.158\textwidth]{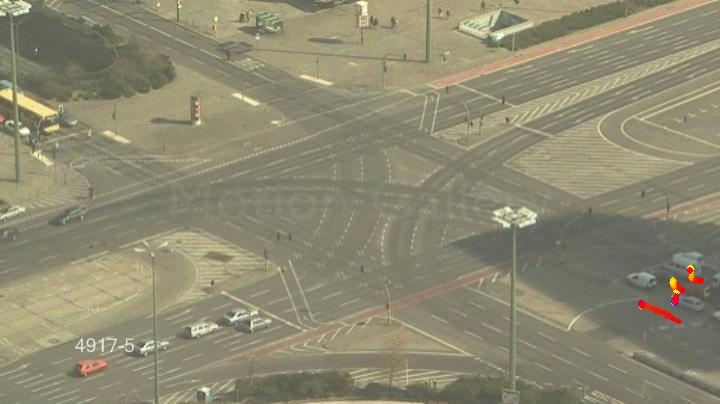}}   
 \subfigure[cluster 11]{\includegraphics[width=0.158\textwidth]{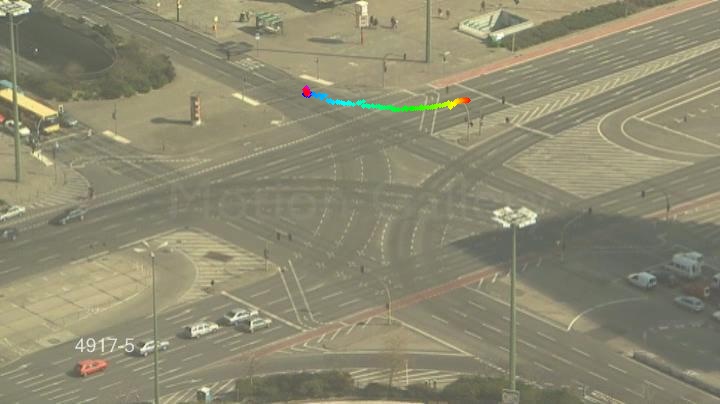}}  
                          
  \caption{Depiction of clustering applied on UCF dataset with TIGM ($\beta = 240$) using all features. (a) All clusters are shown using unique colors. It has been observed that each cluster forms a unique pattern. (b - i) Prominent clusters shown using color gradient form. (j - l) Rare patterns. }
\label{chap4_Fig:UCF}
\end{figure}
\begin{figure}[!h]
  \centering
 \subfigure[Clusters]{\includegraphics[width=0.117\textwidth]{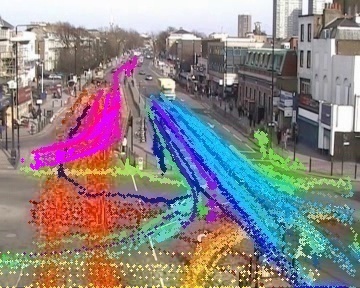}}
 \subfigure[cluster 1]{\includegraphics[width=0.117\textwidth]{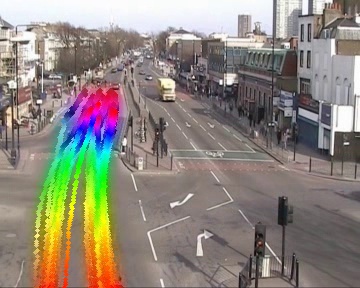}}
 \subfigure[cluster 2]{\includegraphics[width=0.117\textwidth]{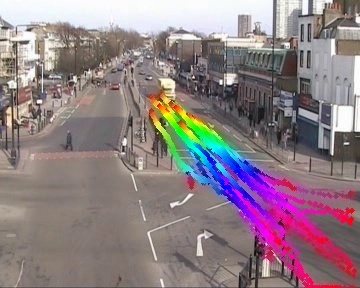}}
 \subfigure[cluster 3]{\includegraphics[width=0.117\textwidth]{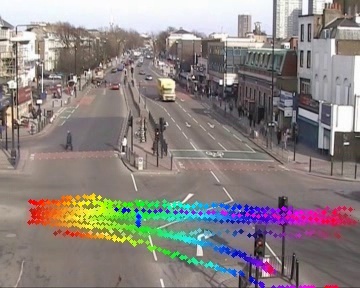}}
 \subfigure[cluster 4]{\includegraphics[width=0.117\textwidth]{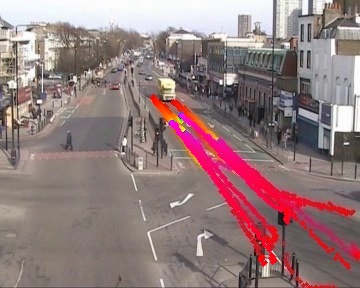}} 
 \subfigure[cluster 5]{\includegraphics[width=0.117\textwidth]{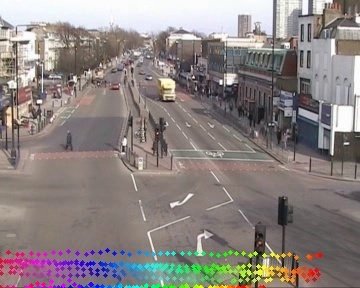}}
 \subfigure[cluster 6]{\includegraphics[width=0.117\textwidth]{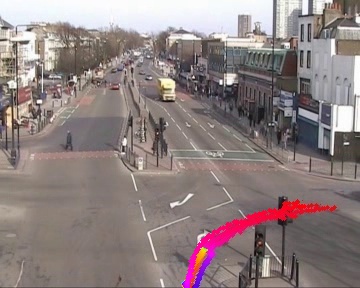}}
 \subfigure[cluster 7]{\includegraphics[width=0.117\textwidth]{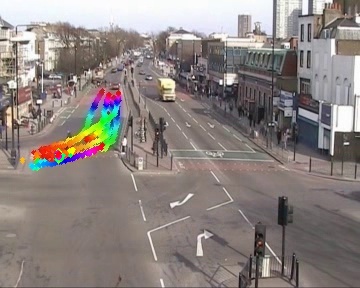}}    
  \caption{Depiction of trajectory clustering applied on QML dataset with TIGM ($\beta = 170$). (a) Clusters shown in unique colors. (b - h) Most prominent clusters  based on the number of tracks shown in color gradient form. }
  \label{chap4_Fig:QML}
\end{figure}
\begin{figure}[!h]
  \centering
 \subfigure[cluster 1]{\includegraphics[width=0.117\textwidth]{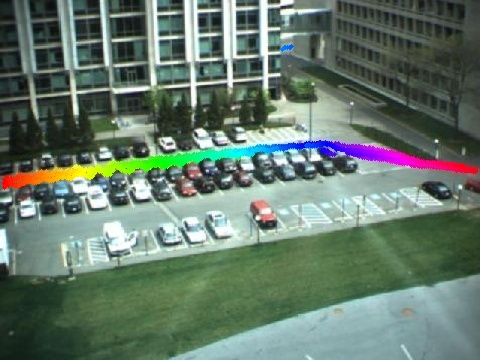}}
 \subfigure[cluster 3]{\includegraphics[width=0.117\textwidth]{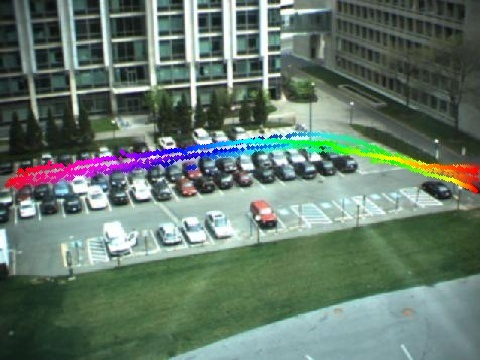}} 
 \subfigure[cluster 4]{\includegraphics[width=0.117\textwidth]{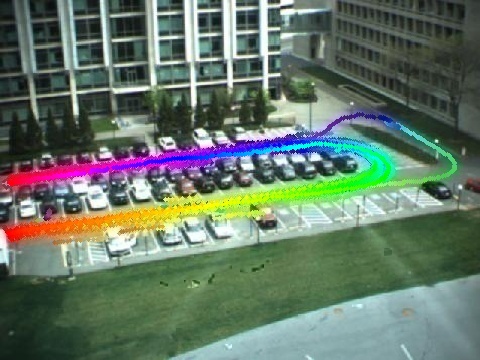}} 
 \subfigure[cluster 5]{\includegraphics[width=0.117\textwidth]{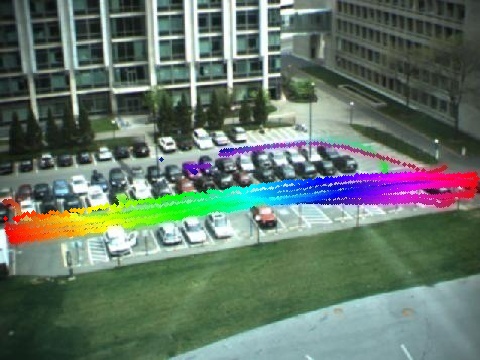}}
 \subfigure[cluster 6]{\includegraphics[width=0.117\textwidth]{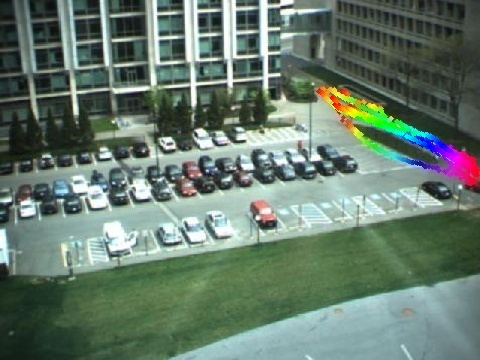}} 
 \subfigure[cluster 7]{\includegraphics[width=0.117\textwidth]{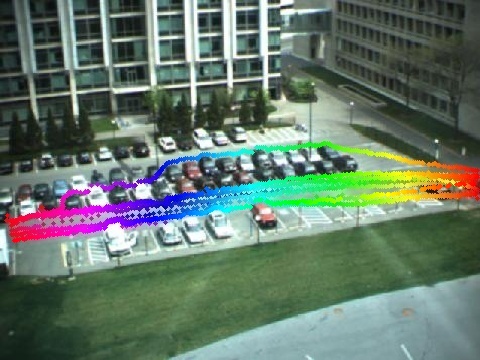}}
 \subfigure[cluster 11]{\includegraphics[width=0.117\textwidth]{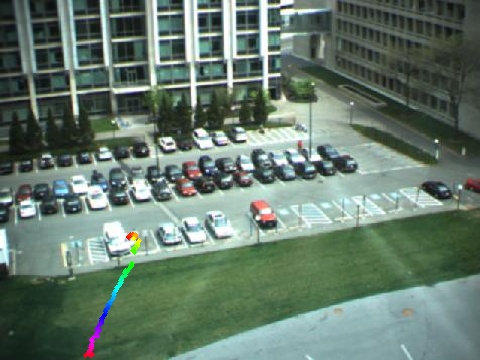}}                       
 \subfigure[cluster 12]{\includegraphics[width=0.117\textwidth]{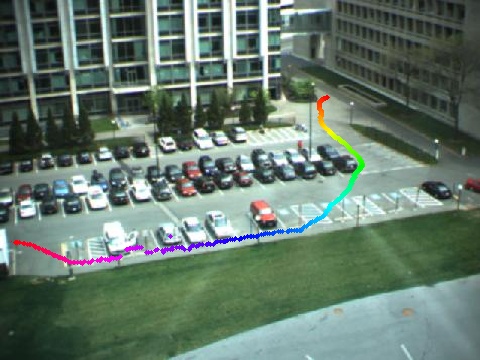}}                                          
  \caption{Depiction of trajectory clustering applied on MIT dataset with TIGM ($\beta = 50$) using start and end positions. (a - f) Prominent clusters are shown in color gradient form. Unlike the UCF and QML datasets, number of clusters formed is large. However, only the top six clusters are shown. (g - h) Less frequent or unusual clusters.}
  \label{chap4_Fig:MIT}
\end{figure}

The clustering results are presented in Figs.~\ref{chap4_Fig:UCF} - \ref{chap4_Fig:MIT}. As expected, $\beta$ variation gives different groupings of tracks. The tracks are incrementally clustered to ensure that subsequent tracks following a similar path are grouped together. This happens due to the neighboring positions of the new tracks with that of the previous one in the multidimensional space. The clusters with more number of tracks form Motifs (frequently occurring trajectory patterns)~\cite{2}. Individual clusters are shown with color gradients indicating the motion direction. Tracks start with red, goes through a change of color and ends with pink to demonstrate the temporal effect. This representation is used to give insight into the motion directions of the trajectories.

\begin{figure}[!h]
  \centering 
 \subfigure[cluster 1]{\includegraphics[width=0.117\textwidth]{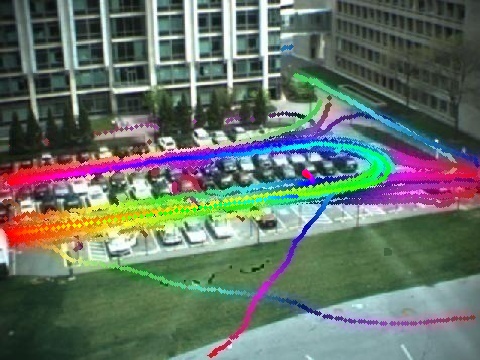}}
 \subfigure[cluster 2]{\includegraphics[width=0.117\textwidth]{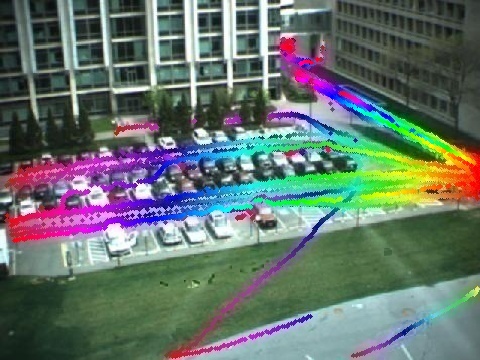}}
 \subfigure[cluster 3]{\includegraphics[width=0.117\textwidth]{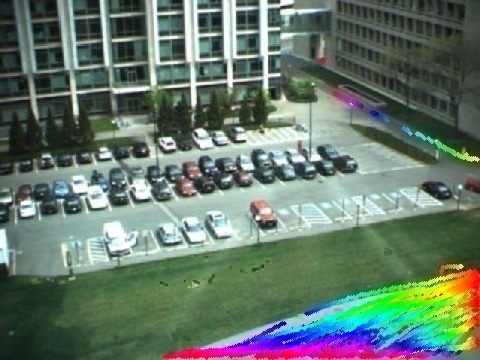}}
 \subfigure[cluster 4]{\includegraphics[width=0.117\textwidth]{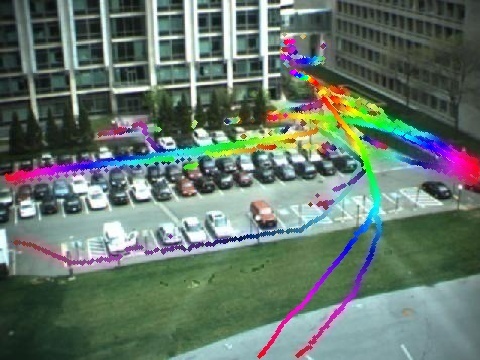}}
 \subfigure[cluster 5]{\includegraphics[width=0.117\textwidth]{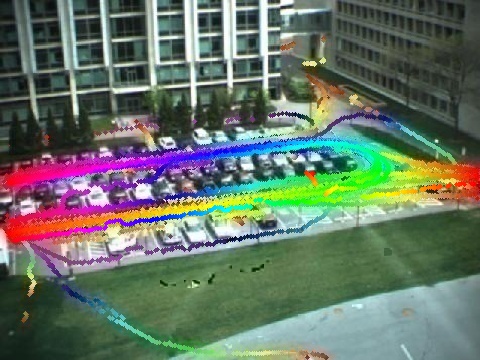}}
 \subfigure[cluster 6]{\includegraphics[width=0.117\textwidth]{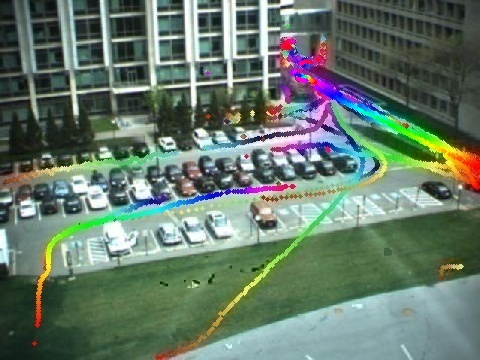}}
 \subfigure[cluster 7]{\includegraphics[width=0.117\textwidth]{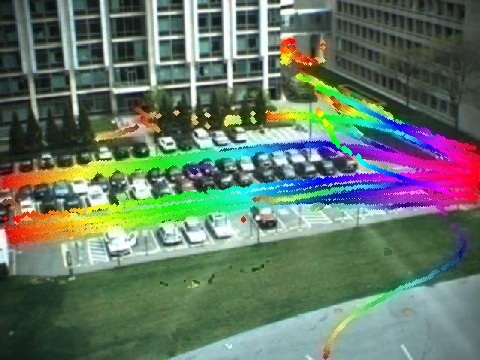}}
 \subfigure[cluster 8]{\includegraphics[width=0.117\textwidth]{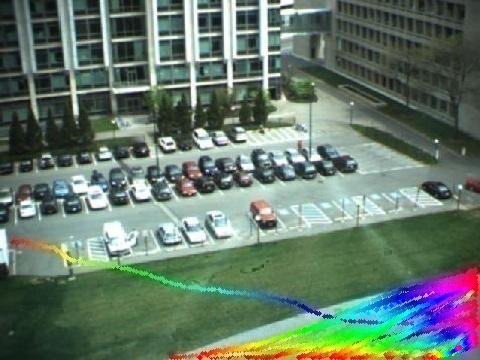}}
  \caption{Depiction of trajectory clustering applied on MIT parking dataset based on start and end points using TIGM ($\beta = 70$). (a - d) Four prominent clusters based on start points. (e - h) Four prominent clusters based on end points.}
  \label{chap4_Fig:MIT2}     
\end{figure}  

In Figs.~\ref{chap4_Fig:UCF}~-~\ref{chap4_Fig:QML}, clustering results considering all five dimensions of the tracks are presented. The results clearly show that trajectories in a cluster are similar in nature. Even though the patterns shown in Fig.~\ref{chap4_Fig:QML}(c) and Fig.~\ref{chap4_Fig:QML}(e) look similar, their time spans are different. The video reveals that Fig.~\ref{chap4_Fig:QML}(c) is formed due to continuous moving of traffic, while Fig.~\ref{chap4_Fig:QML}(e) is formed as a result of the vehicles being stopped at the signal for some duration. It has been observed that the number of clusters formed in MIT dataset is high. This is due to a large number of entry and exit points in the parking area, where many vehicles did not strictly follow the paths while moving in or out. If there are no restrictions on the paths to be taken by the drivers, clustering may be performed based on entry or exit points to localize most frequently used entry and exit points, as given in Fig. \ref{chap4_Fig:MIT2}.

\subsubsection{TIGM Experiments and Analysis on Crowd Data}
\begin{figure}[!h]
  \centering
 \subfigure[cluster 1]{\includegraphics[width=0.117\textwidth]{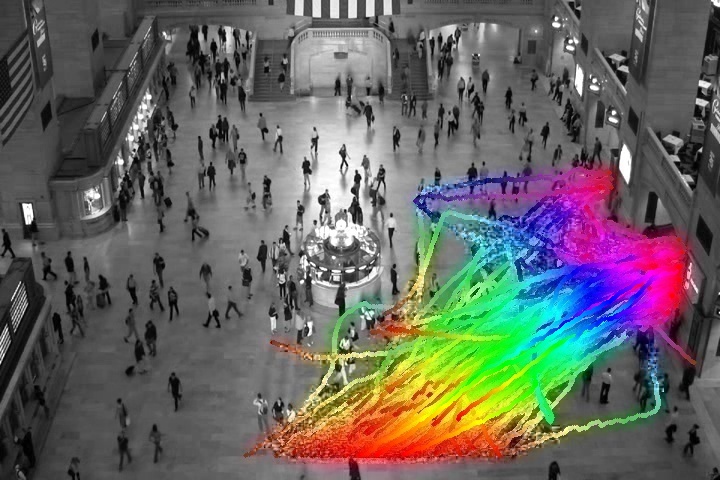}}
 \subfigure[cluster 2]{\includegraphics[width=0.117\textwidth]{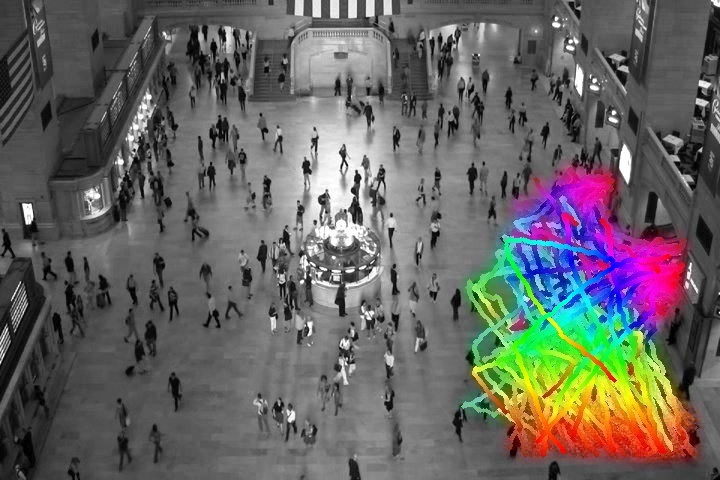}} 
 \subfigure[cluster 3]{\includegraphics[width=0.117\textwidth]{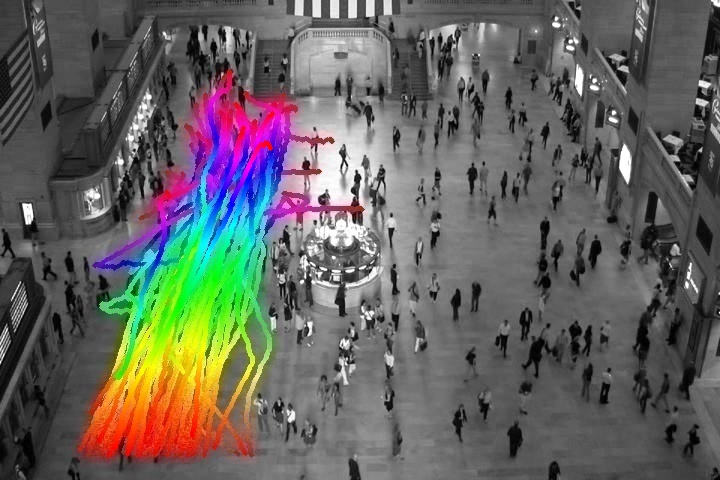}}    
 \subfigure[cluster 4]{\includegraphics[width=0.117\textwidth]{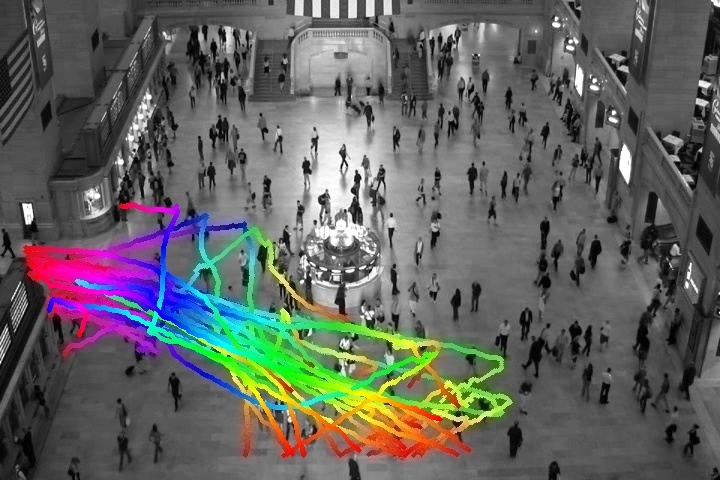}} 
 \subfigure[cluster 5]{\includegraphics[width=0.117\textwidth]{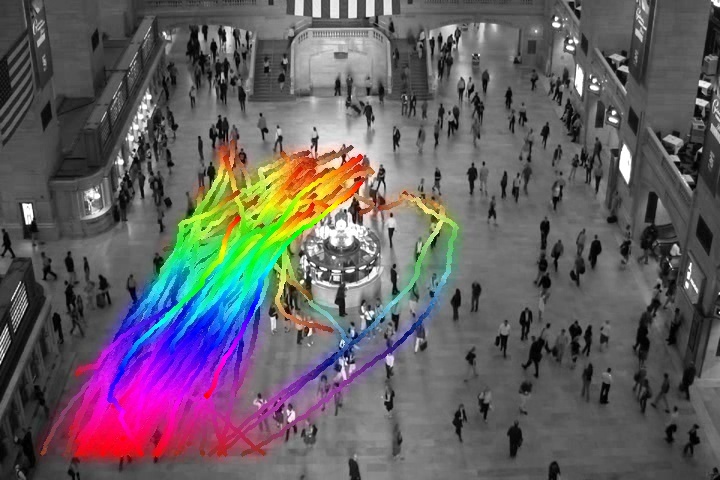}} 
 \subfigure[cluster 6]{\includegraphics[width=0.117\textwidth]{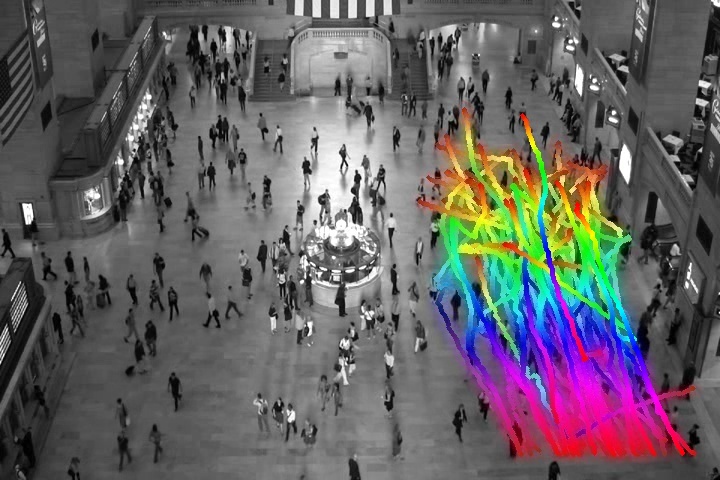}}    
 \subfigure[cluster 7]{\includegraphics[width=0.117\textwidth]{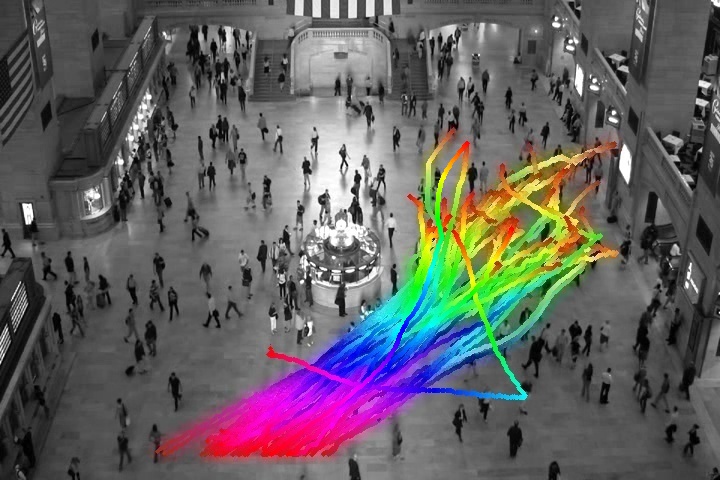}}  
 \subfigure[cluster 8]{\includegraphics[width=0.117\textwidth]{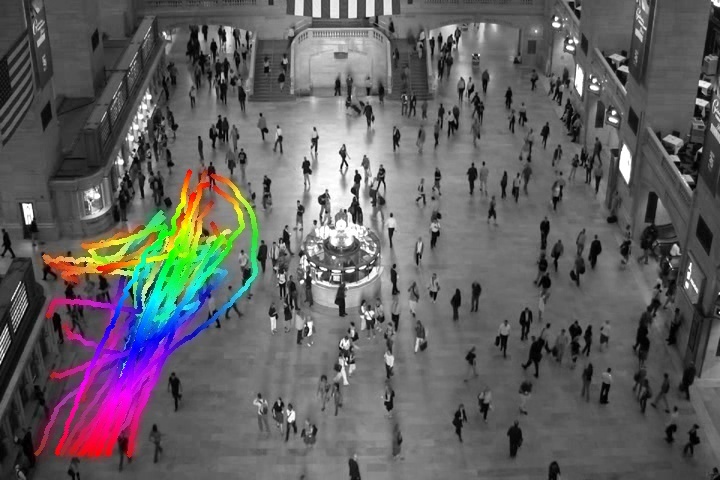}}   
 \subfigure[cluster 9]{\includegraphics[width=0.117\textwidth]{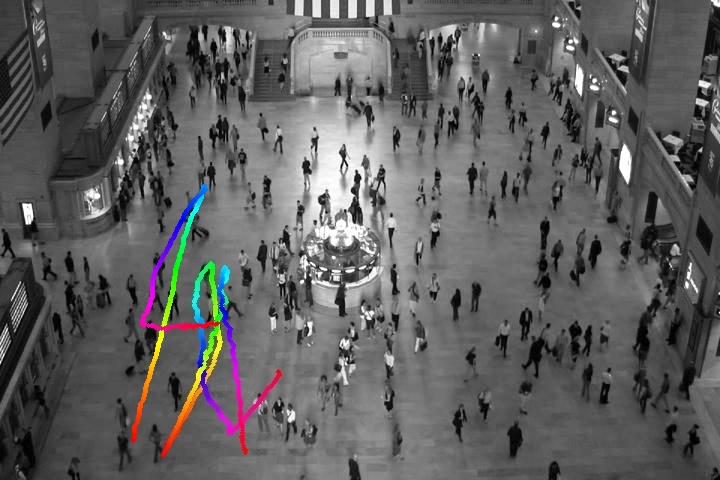}}  
 \subfigure[cluster 10]{\includegraphics[width=0.117\textwidth]{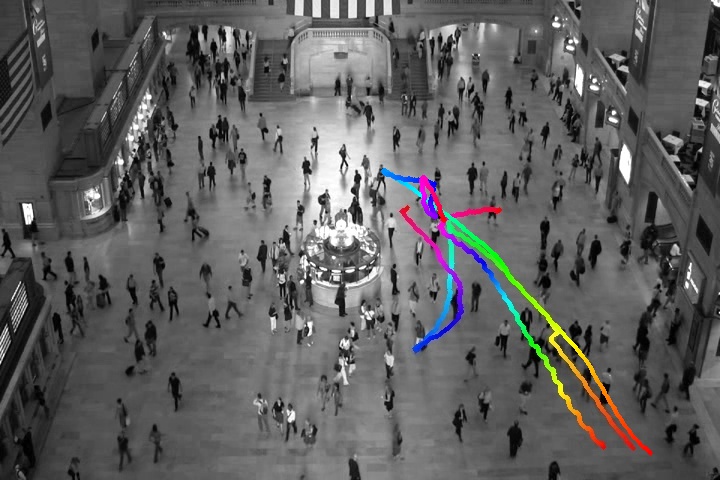}}  
 \subfigure[cluster 11]{\includegraphics[width=0.117\textwidth]{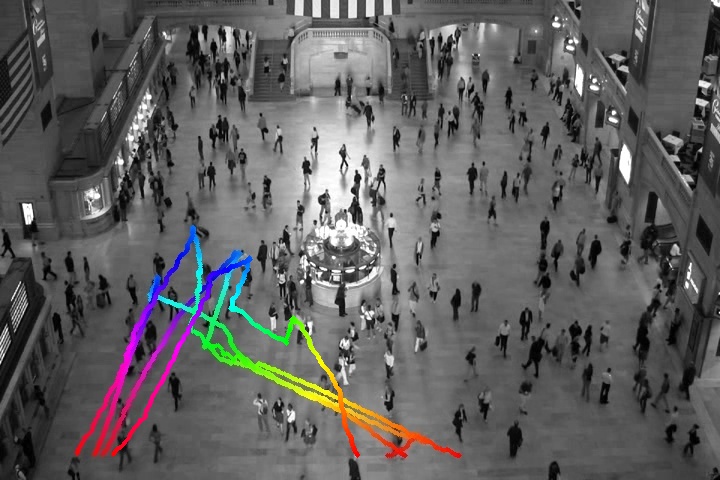}}   
 \subfigure[cluster 12]{\includegraphics[width=0.117\textwidth]{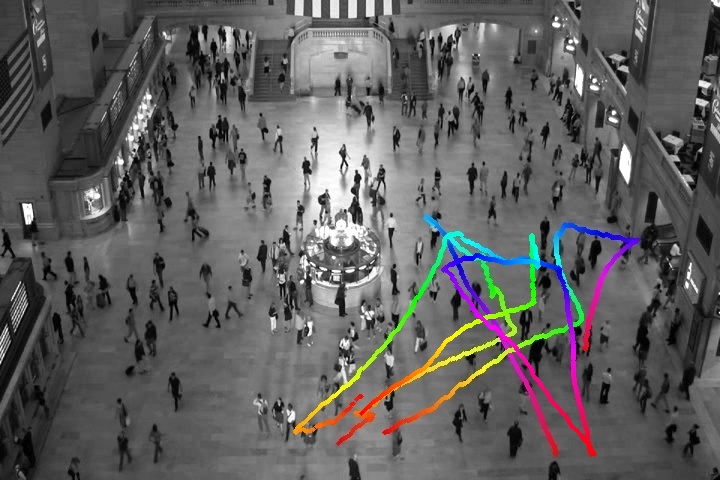}}  
 \subfigure[cluster 13]{\includegraphics[width=0.24\textwidth]{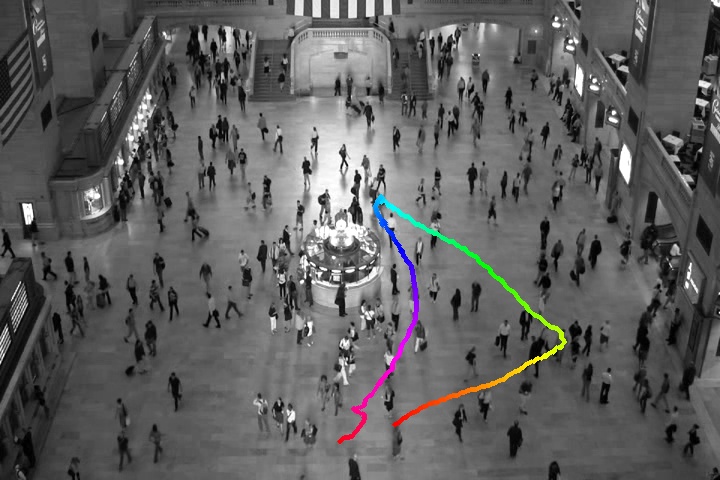}}  
 \subfigure[cluster 14]{\includegraphics[width=0.24\textwidth]{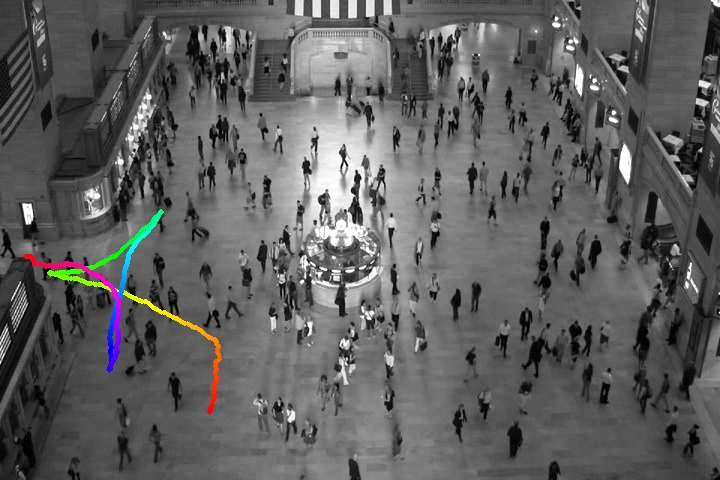}}  
  \caption{Depiction of trajectory clustering applied on GCS dataset using TIGM ($\beta = 170$) using all 5 dimensions.  (a - h) Clusters representing the noticeable patterns. (i - l) Clusters representing less frequently occurring patterns. (m - n) Clusters representing unusual trajectories. }
  \label{chap4_Fig:STATION}
\end{figure}
\begin{figure}[!h]
  \centering
 \subfigure[cluster 1]{\includegraphics[width=0.117\textwidth]{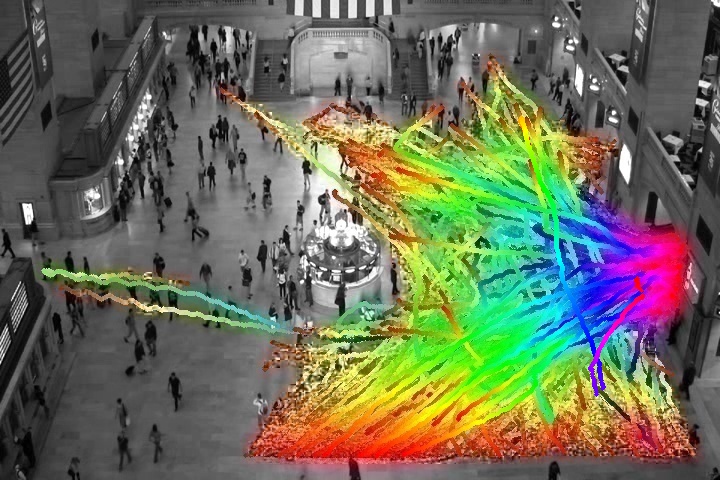}}
 \subfigure[cluster 2]{\includegraphics[width=0.117\textwidth]{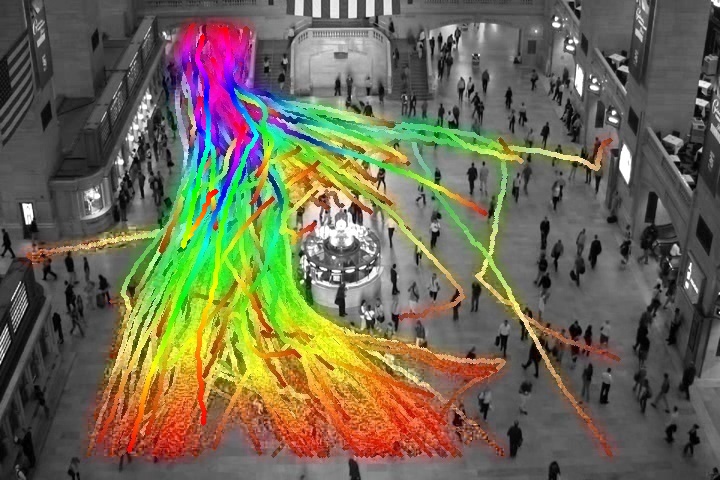}} 
 \subfigure[cluster 3]{\includegraphics[width=0.117\textwidth]{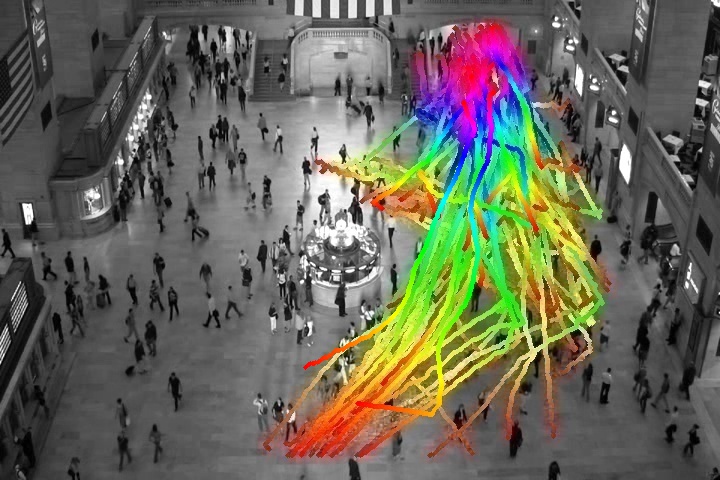}} 
 \subfigure[cluster 4]{\includegraphics[width=0.117\textwidth]{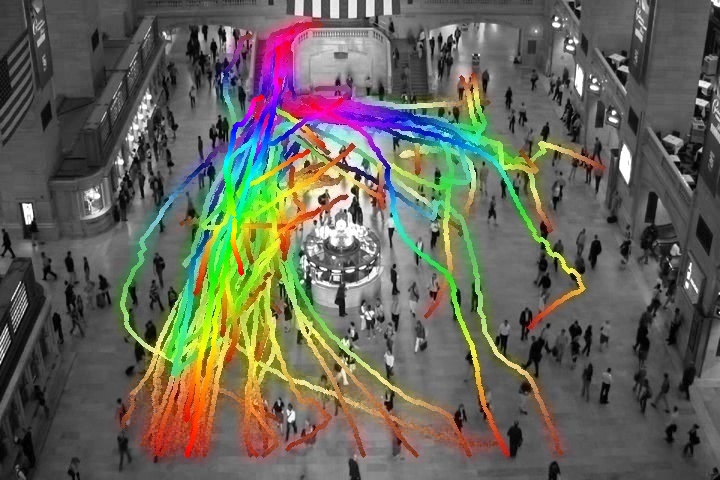}} 
 \subfigure[cluster 5]{\includegraphics[width=0.117\textwidth]{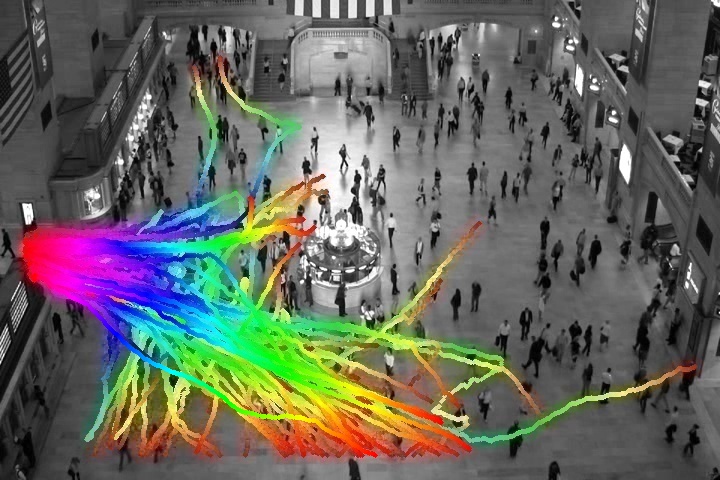}} 
 \subfigure[cluster 6]{\includegraphics[width=0.117\textwidth]{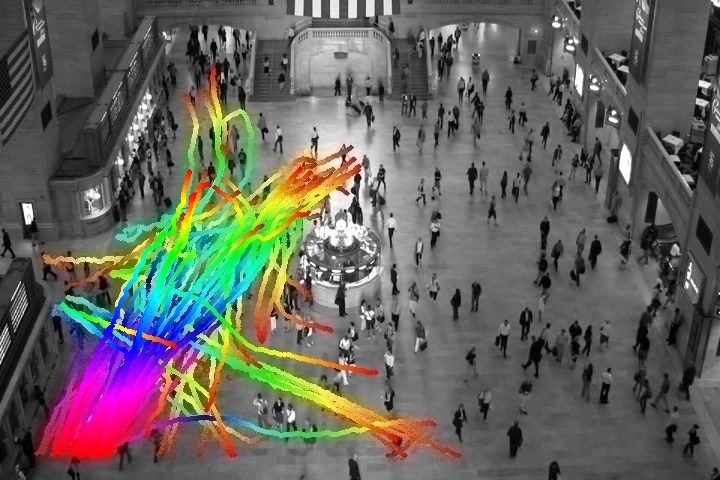}}     
 \subfigure[cluster 7]{\includegraphics[width=0.117\textwidth]{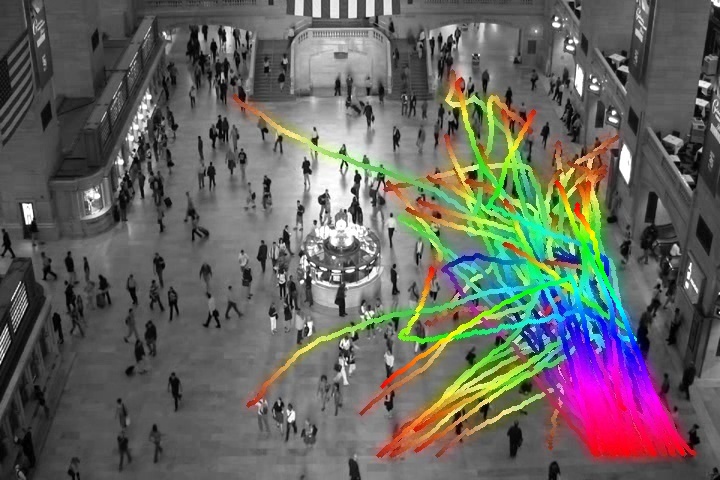}}  
 \subfigure[cluster 8]{\includegraphics[width=0.117\textwidth]{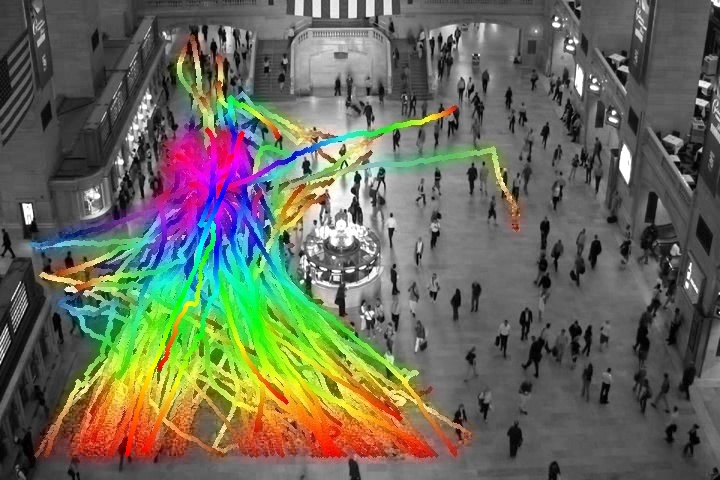}}    
  \caption{Depiction of Trajectory clustering applied on GCS dataset based on destination using TIGM ($\beta = 70$). (a - h) Prominent clusters. Clusters clearly indicate the exit paths as well as the place, where people may stop mid-course.}
  \label{chap4_Fig:STATION_DST}
\end{figure}
\begin{figure}[!h]
  \centering
 \subfigure[cluster 1]{\includegraphics[width=0.117\textwidth]{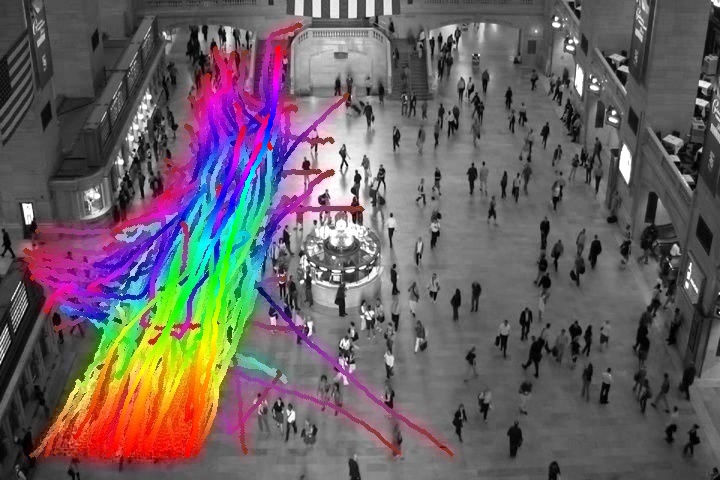}} 
 \subfigure[cluster 2]{\includegraphics[width=0.117\textwidth]{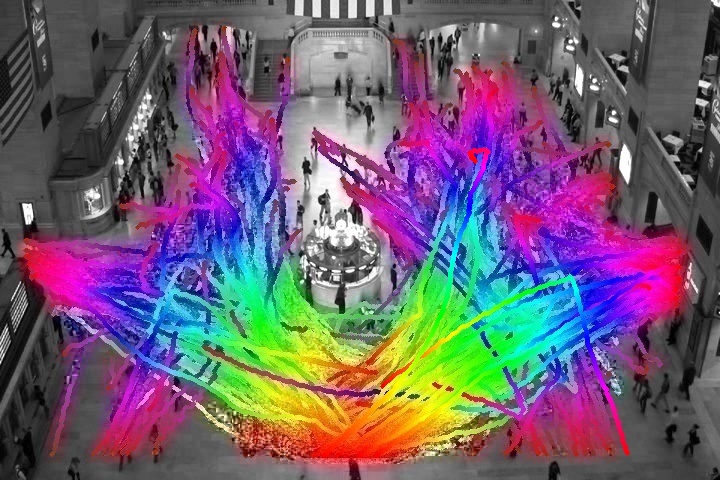}}   
 \subfigure[cluster 3]{\includegraphics[width=0.117\textwidth]{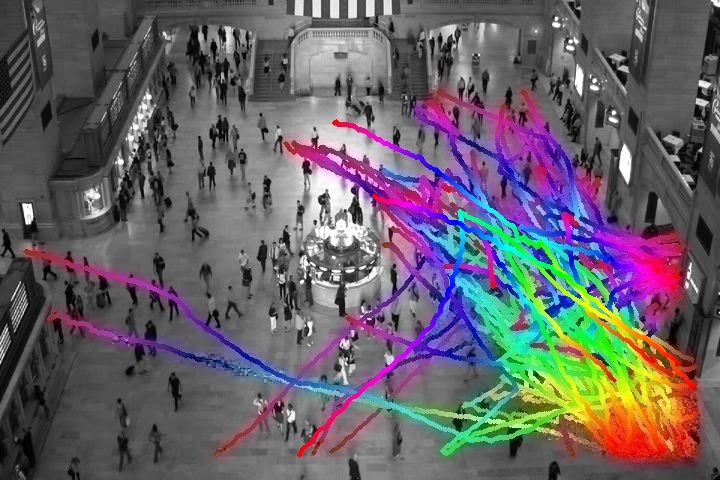}} 
 \subfigure[cluster 4]{\includegraphics[width=0.117\textwidth]{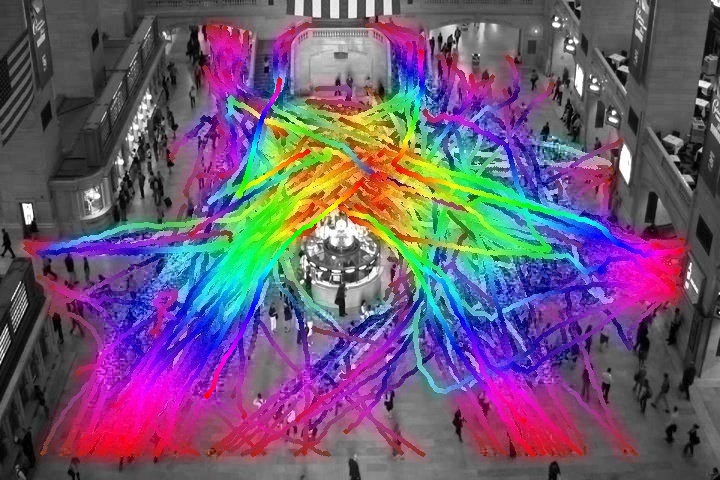}}   
 \subfigure[cluster 5]{\includegraphics[width=0.117\textwidth]{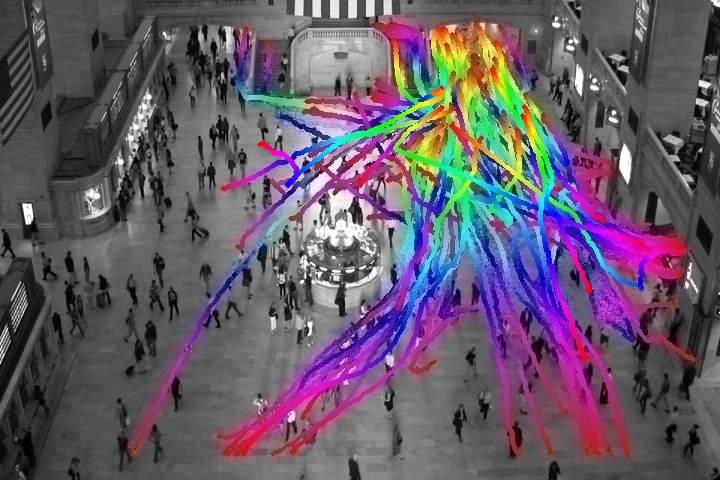}}     
 \subfigure[cluster 6]{\includegraphics[width=0.117\textwidth]{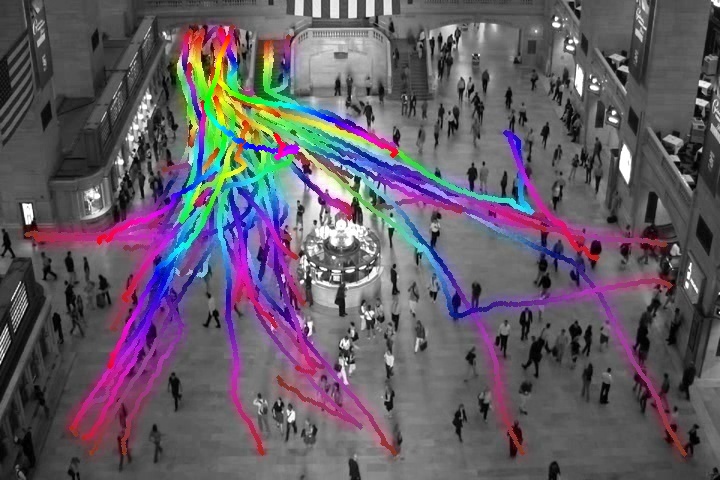}} 
  \subfigure[cluster 7]{\includegraphics[width=0.117\textwidth]{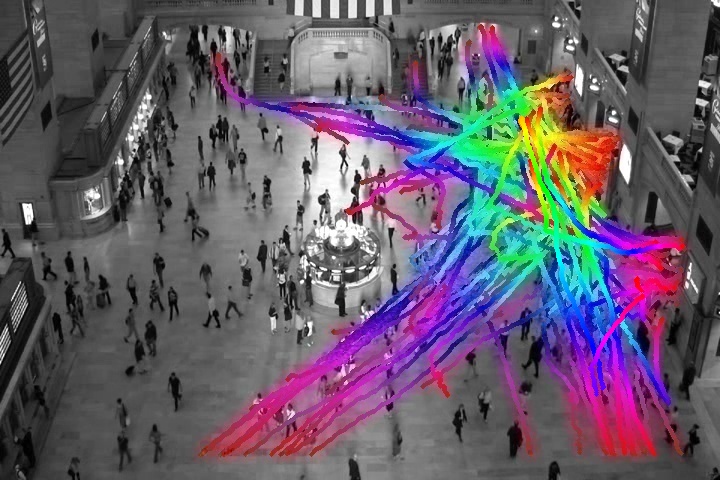}} 
 \subfigure[cluster 8]{\includegraphics[width=0.117\textwidth]{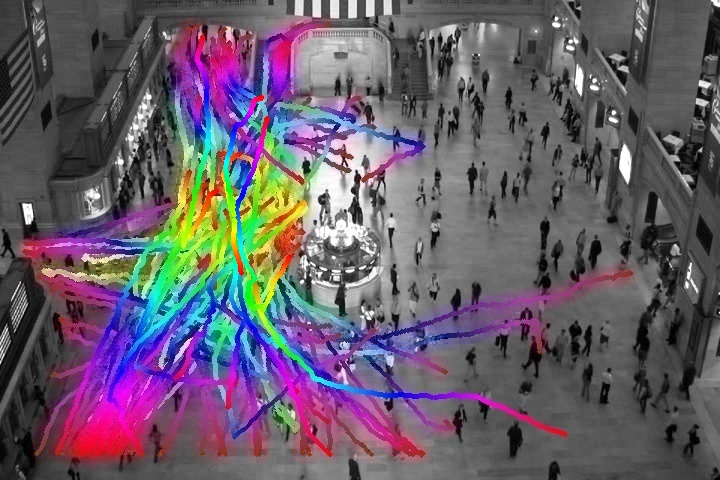}}
  \caption{Depiction of Trajectory clustering applied on GCS dataset based on source of trajectories using TIGM ($\beta = 100$). (a - h) Prominent Clusters. Clusters clearly indicate the entry paths as well as the place, where  people usually resume their movement.}
  \label{chap4_Fig:STATION_SRC}
\end{figure}
Applicability of the model on crowd data has been tested to understand the underlying patterns of crowd movements. Experiments on a challenging dataset like GCS reveal that the proposed method can produce interesting results, including detection of frequently used paths, unusual patterns of people movement, most frequently used entry or exit points, etc.  Tracks of duration larger than $400$ frames have been used to avoid the truncated trajectories. Clustering results, as demonstrated in Figs.~\ref{chap4_Fig:STATION}~-~\ref{chap4_Fig:STATION_SRC} reveal interesting patterns including most frequently used entry and exit points. In Fig. \ref{chap4_Fig:STATION}, the clusters obtained considering all five dimensions are presented. With the increasing number of trajectories in a cluster, the pattern becomes more prominent as visible in Fig.~\ref{chap4_Fig:STATION}. It also indicates whether the pattern is dense or less frequently appearing. The clusters with lesser trajectories indicate anomalous or rare behavior. Since the space covered in the scene does not restrict movement of people, start or end points guided clustering can reveal important information about the entry and exit points in the scene as shown in Figs.~\ref{chap4_Fig:STATION_DST}~-~\ref{chap4_Fig:STATION_SRC}.
\subsection{DEM Experiments and Result Analysis}
\label{chap4_sec:DEM}
  In order to conduct experiments to detect scene dynamics, GCS (crowd) and QML (traffic) datasets have been used. For GCS, the inference is applied by taking a time segment ($\Delta t$) of $10$K frames for the DEM. For the QML dataset, 15K frames per time segment have been used.
\subsubsection{Analysis of GCS (crowd) Results}
\begin{figure}[!h]
  \centering
      \includegraphics[scale = 0.3]{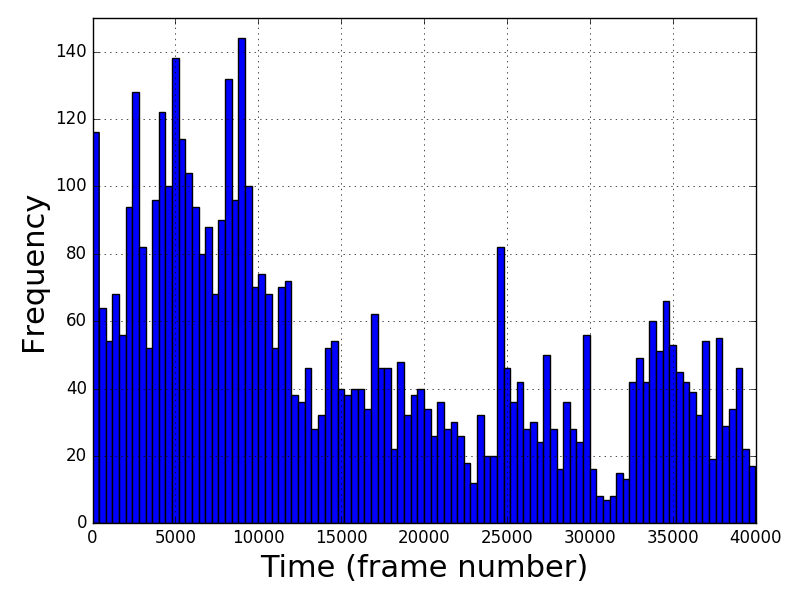}
    \caption{Histogram depicts the number of new tracks against their arrival time in GCS. }
  \label{chap4_Fig:TempTracks}  
\end{figure} 
\begin{figure*}[!h]
  \centering
\subfigure[Top-left entry]{\includegraphics[width=0.19\textwidth]{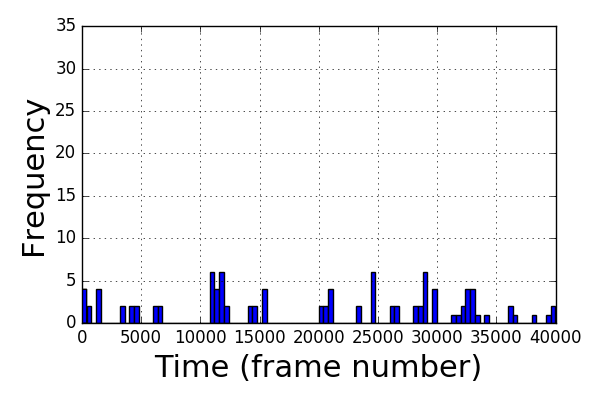}}    
\subfigure[Dynamics of cluster 6]{\includegraphics[width=0.79\textwidth]{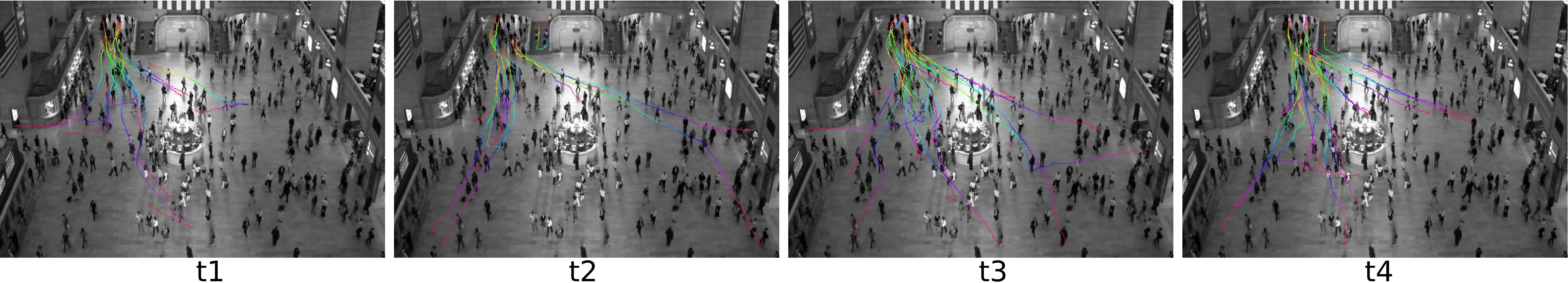}}  
\subfigure[Middle-bottom entry]{\includegraphics[width=0.19\textwidth]{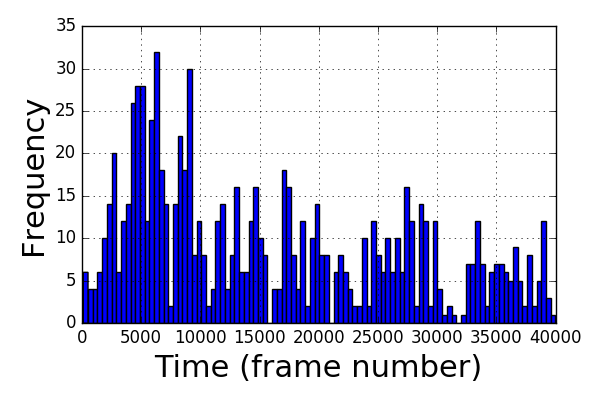}}  
\subfigure[Dynamics of cluster 9]{\includegraphics[width=0.79\textwidth]{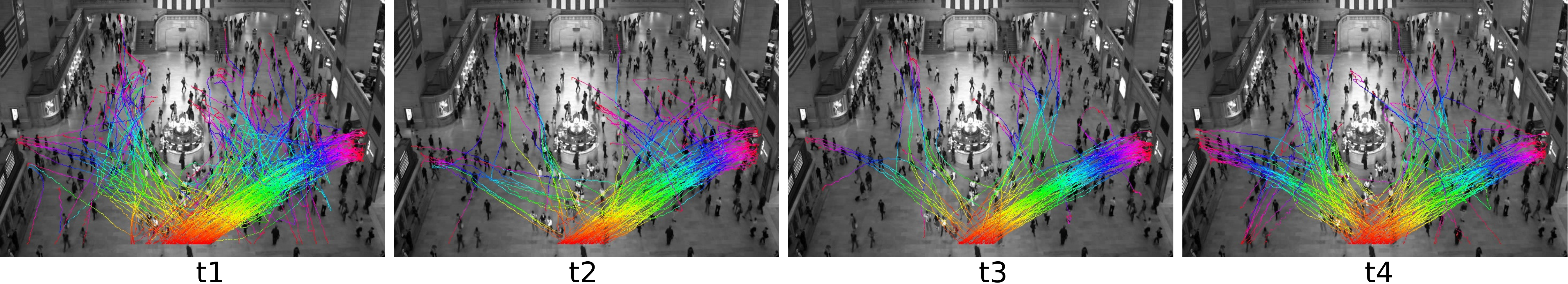}}  
  \caption{A depiction of pattern dynamics as represented by clusters (with $\beta = 100$) at four equally spaced time segments for top-left and middle-bottom entry areas in the GCS dataset. Here, (a) and (c) represent the trajectory arrival frequencies for top-left and middle-bottom entry areas. (b) At the top-left entry area, the crowd density is spare and steady over  all time segments. (d) Initially, crowd in slightly higher and then becomes steady at the middle-bottom entry area.}
  \label{chap4_Fig:DEM_GRAND}
\end{figure*}
  In order to illustrate how DEM captures crowd dynamics,  consider the graph depicting frequency of tracks vs. appearance time as shown in Fig.~\ref{chap4_Fig:TempTracks}. Since the frequency is an indication of arrival of people at the scene, this plot suggests that initially there is a higher crowd density, which reduces with time. However, this does not speak much  about the arrival events at different regions of the scene. For example, consider the frequency plots of top-left and middle-bottom entry areas, as shown in Fig.~\ref{chap4_Fig:DEM_GRAND}(a) and Fig.~\ref{chap4_Fig:DEM_GRAND}(c). Start point-based DEM clustering results shown in Fig.~\ref{chap4_Fig:DEM_GRAND}(b) and Fig.~\ref{chap4_Fig:DEM_GRAND}(d) reveal that the arrival trend is not same for two regions. Here, the dynamics reflects the crowd density information over time. This information can be used for crowd management. For example, deciding the number of counters for issuing tickets or scheduling of trains can be done based on change in density.
Though start point-based clustering on DEM has been used in the experiment, this can be extended by combining other features.
\subsubsection{Analysis of QML (traffic) Results} 
\begin{figure}[!h]
  \centering
\subfigure[Dynamic clustering results using start and end points with $\beta = 70$]{\includegraphics[width=0.49\textwidth]{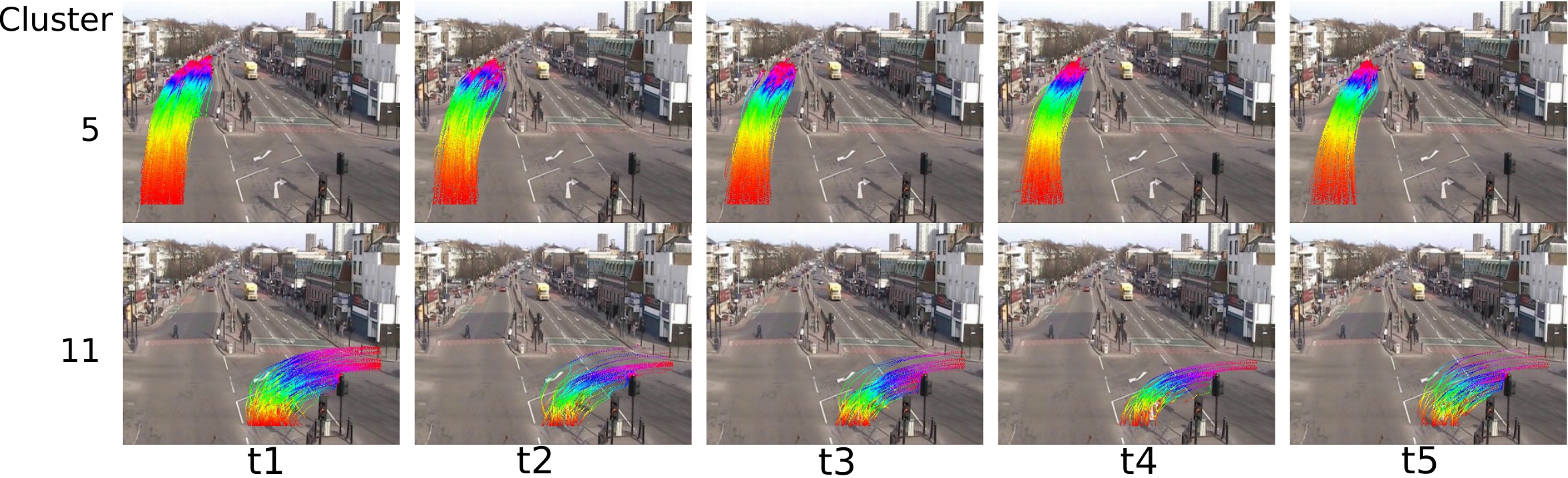}}
\subfigure[Dynamic clustering results using end points with $\beta = 50$]{\includegraphics[width=0.49\textwidth]{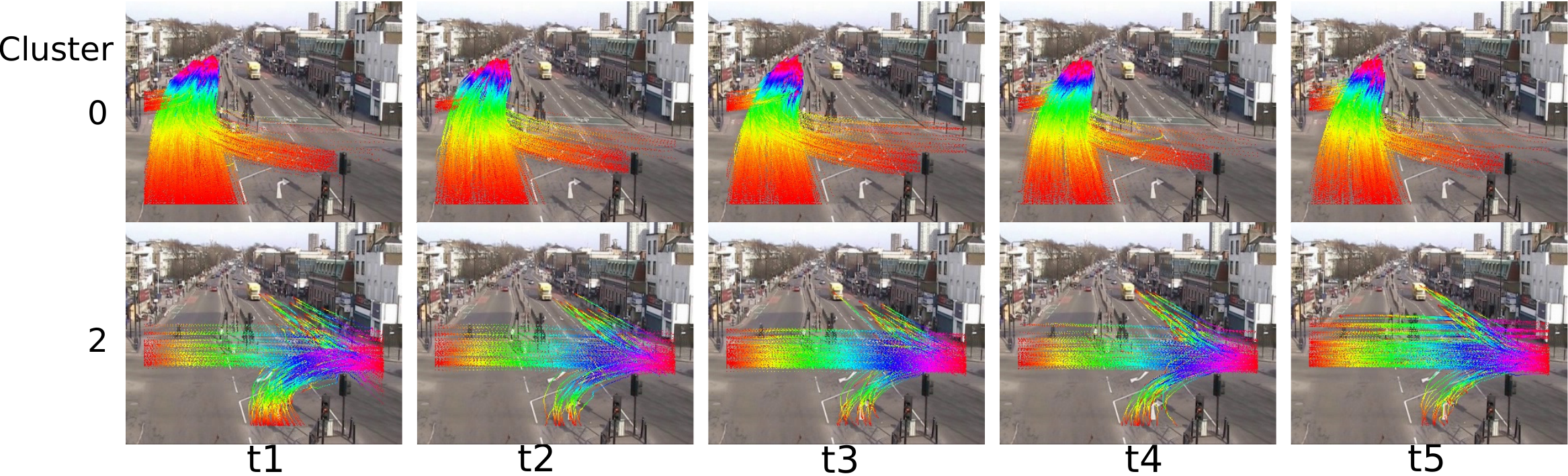}}
\subfigure[Density plot for (a)]{\includegraphics[width=0.24\textwidth]{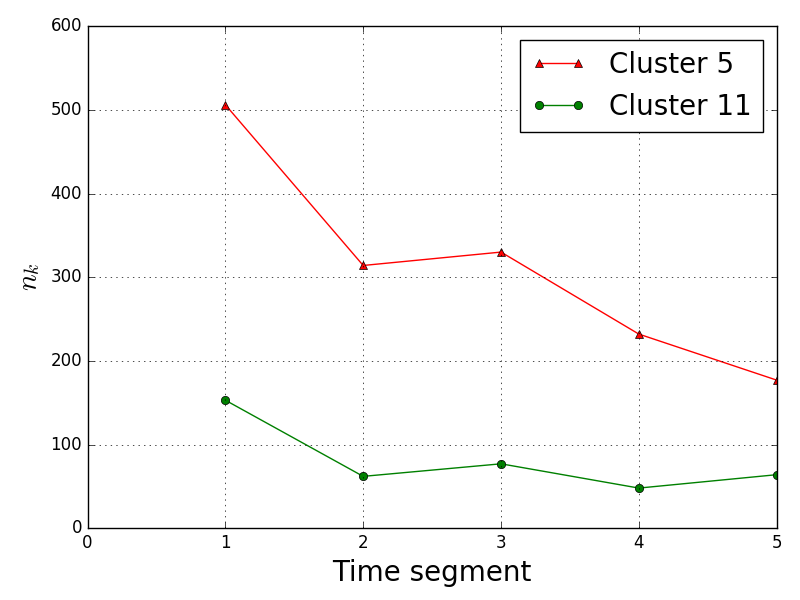}}
\subfigure[Density plot for (b)]{\includegraphics[width=0.24\textwidth]{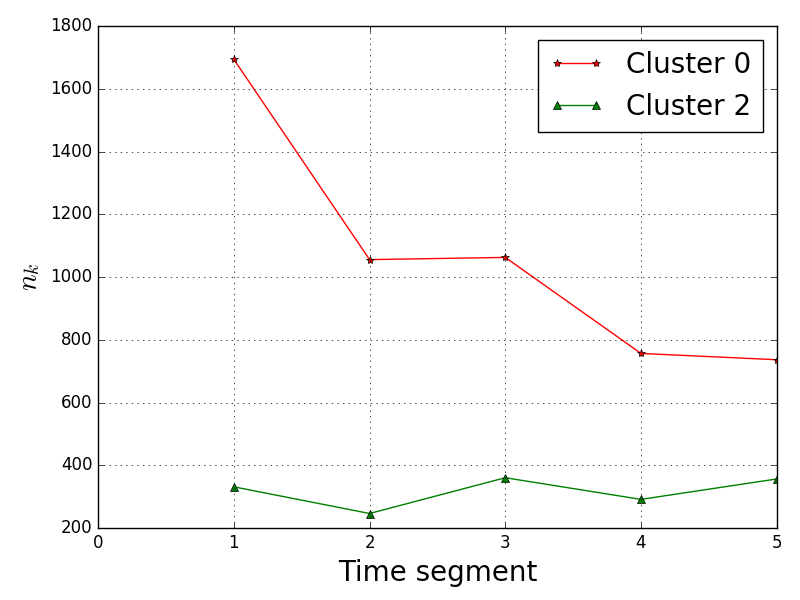}}
  \caption{A depiction of DEM on QML dataset on selected clusters. Here, (a) \& (b) show the visualization of the clusters and (c) \& (d) show the respective cluster size variation over time.  In (a), row 1 represents south-to-north traffic and row 2 represents south-to-east traffic.  In (b), row 1 represents the north-bound traffic and row 2 represents the east-bound traffic.}
  \label{chap4_Fig:DEM_QML_SRC_DST_BETA_70}
\end{figure}
\begin{figure*}[h!]
  \centering
\subfigure[Traffic scenarios]{\includegraphics[width=0.35\textwidth]{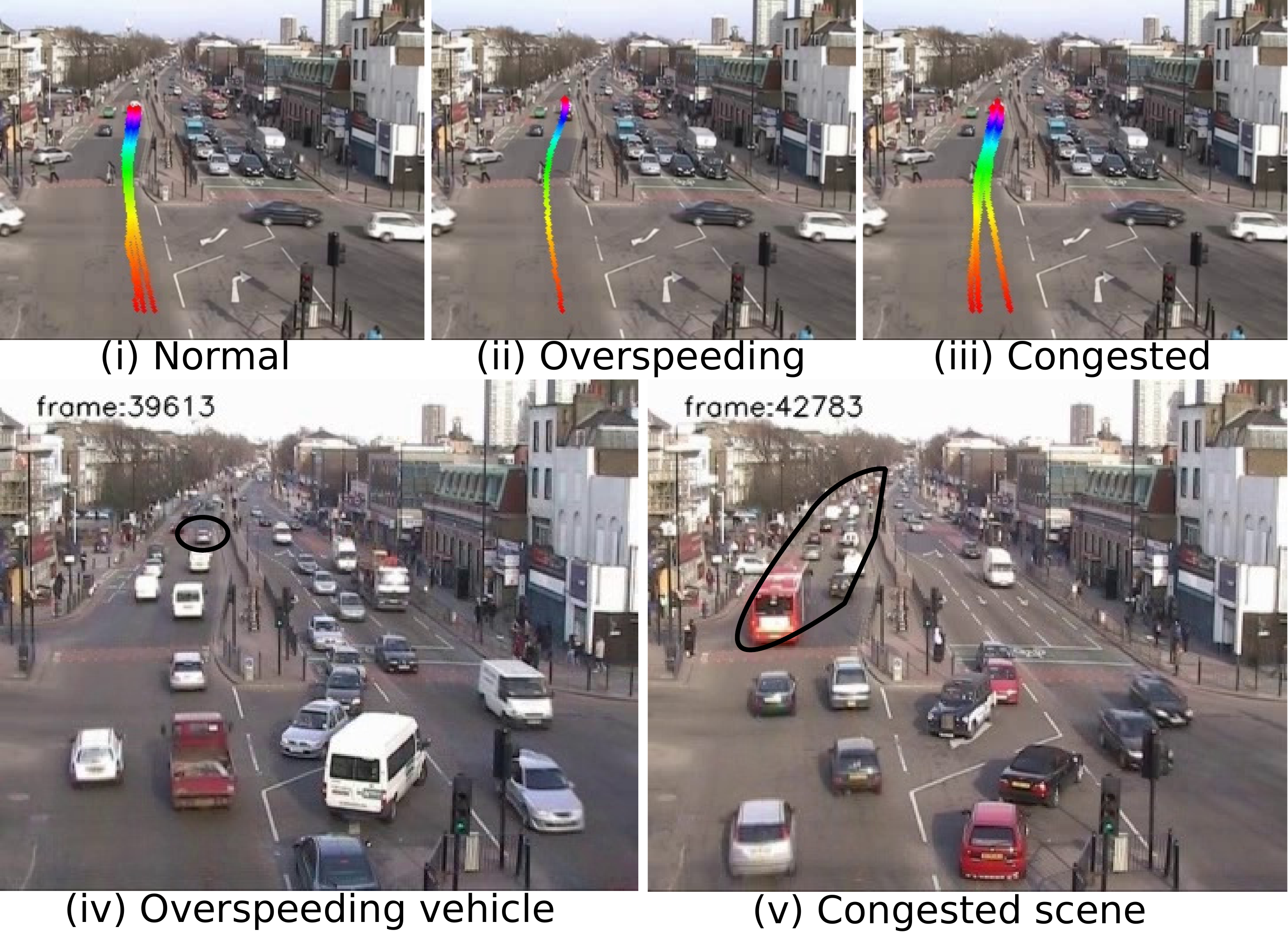}}
\subfigure[Duration plot]{\includegraphics[width=0.36\textwidth]{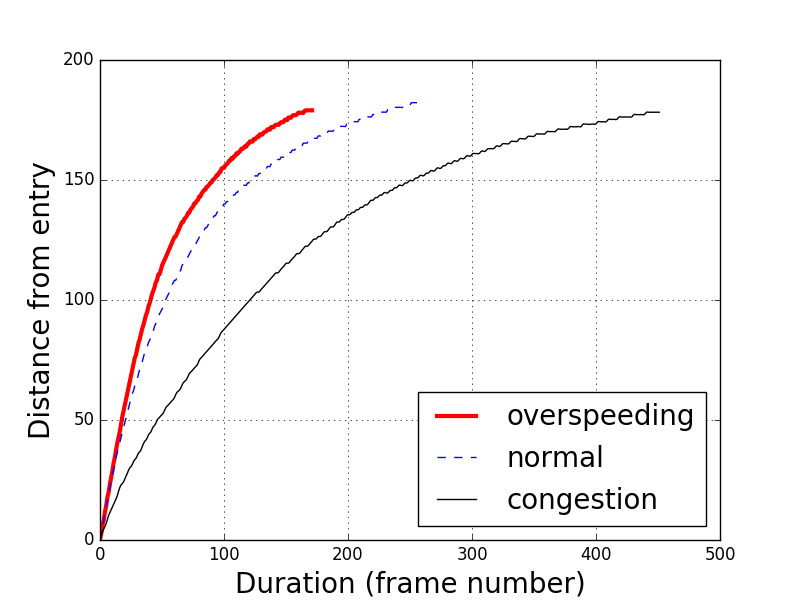}}
\subfigure[Illustration]{\includegraphics[width=0.27\textwidth]{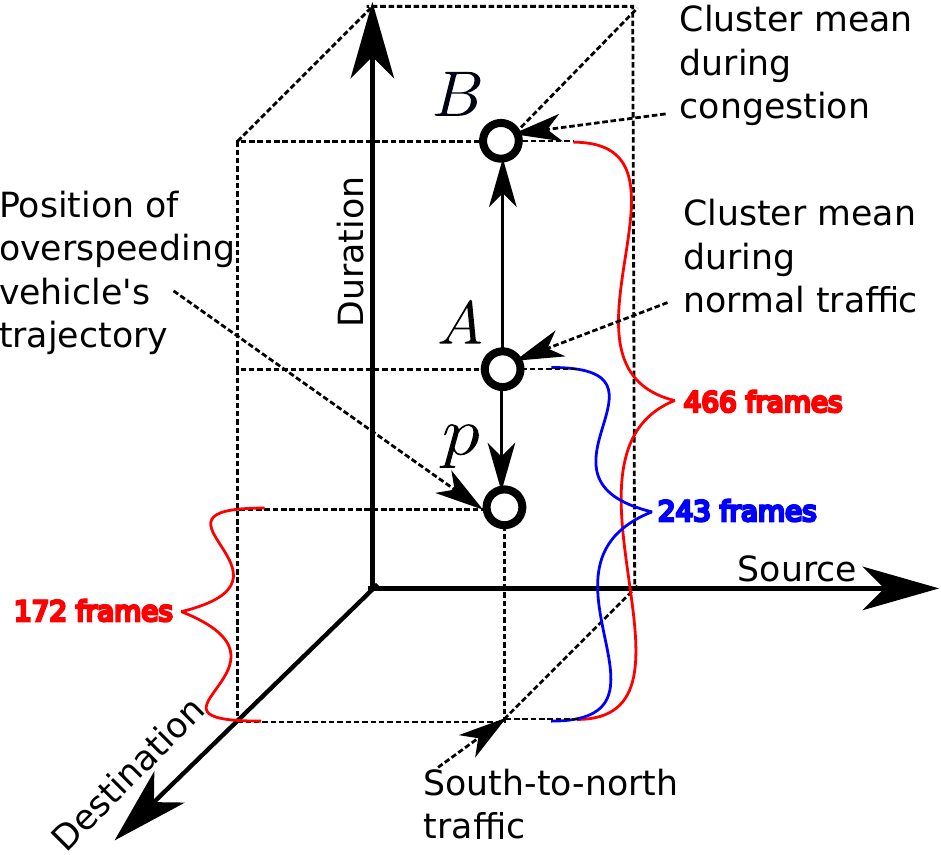}}
  \caption{An illustration of how the DEM captures congestion and overspeeding in the right lane of south-to-north traffic. DEM clustering (using $\beta = 70$ with start and end point features) has been done with $\Delta t = 3600$ frames ($2$ minutes). In the  segment between $21^{st}$ and $23^{rd}$ minutes, vehicles move smoothly representing normal scene. In the next time segment, the traffic slows down due to possible congestion. (a) The first row represents the trajectories in color gradient. The second row represents the corresponding traffic scenes. In the time segment between $21^{st}$ and $23^{rd}$ minutes, the average trajectory duration is found to be 243 frames. However in the next time segment, there is a congestion that makes the average trajectory duration approximately 466 frames. (b) Plot showing instantaneous distance of a point on a sample trajectory from the start location (north-bound traffic) in conditions such as  normal, overspeeding, and probable congestion. It may be observed that the time to cover approximately same distance is significantly different for the three scenarios. (c) Visualization  of the respective scenarios represented using three axes corresponding to source, destination, and duration of the trajectories. Point $A$ denotes the cluster mean of normal trajectories. Movement of the cluster mean ($\mu_k$) from $A$ to $B$ indicates a  possible congestion. When an object moves from a source to destination, a trajectory duration can become longer if the traffic is   moving slowly.  The point $p$ corresponds to an overspeeding vehicle's trajectory, where the track duration is significantly lesser than the average trajectory at normal conditions.}
  \label{chap4_Fig:Analysis}
\end{figure*} 

 The results of the experiments are presented in Fig. \ref{chap4_Fig:DEM_QML_SRC_DST_BETA_70}. It shows how traffic flow changes over time at different paths. For example, cluster 5 as shown in Fig. \ref{chap4_Fig:DEM_QML_SRC_DST_BETA_70}(a) represents a path from south to north, where the traffic volume reduces over a span of approximately 9 minutes as depicted in Fig.~\ref{chap4_Fig:DEM_QML_SRC_DST_BETA_70}(c).  However, as density is relatively higher, green signal duration can be adjusted for this path.   Similarly, cluster 11 depicts the path for south-to-east bound traffic. It may be observed from Fig.~\ref{chap4_Fig:DEM_QML_SRC_DST_BETA_70}(c) that the density decreases gradually and becomes steady as the time slots progress. This signifies that the model is able to capture the change in traffic dynamics along various paths. Similar characteristics can be observed for end point-based DEM clustering as shown in Fig.~\ref{chap4_Fig:DEM_QML_SRC_DST_BETA_70}(b) and Fig.~\ref{chap4_Fig:DEM_QML_SRC_DST_BETA_70}(d). Higher traffic flow can be observed consistently for the north-bound route as compared to east-bound route. Though the north-bound traffic steadily decreases with time, the east-bound traffic is more or less steady. The information can be used for coordinated signaling in a given road network.
 
\subsubsection{DEM for Traffic Applications}  
 Fig.~\ref{chap4_Fig:Analysis} gives an insight into the traffic congestion and anomaly detection capabilities of the DEM. For traffic signaling applications, features can be used in three ways: (i) start and end points-based (reflecting the traffic volume in a path), (ii) only start point-based (reflecting traffic volume from an entry point), and (iii) only end point-based (reflecting the traffic volume towards an exit).  Since the model reflects the change in traffic trend, it can be used for adaptive  timing of signals. For example, cluster density in a path can be used for signal duration prediction, while traffic volume information at the start or end points can be used for better routing at the preceding or succeeding traffic junctions with suitable communication through centralized/ Internet of Things (IoT) devices. History analysis of traffic parameter ($\Phi_k^t$) over a period of days (e.g. one month) can reflect the general traffic trend on different days of a week. The data can also be used for infrastructure expansions if bottlenecks (e.g. frequent congestions) are found.
 
\subsection{Complexity Analysis}
 When $k$ clusters are obtained from $n$ observations, TIGM clustering complexity is $\Theta(nk)$ as each observation needs to be compared with exactly once with the existing $k$ clusters. Even with  DEM, its complexity does not change. The proposed algorithm takes only one observation at a time and the inference process is applied to obtain the clusters. At the time of unassigning the older observations from the cluster, exactly one more revisit is required. Each of the observations is checked only once against each of the $k$ clusters. If resampling is involved, it is done at most twice.  Thus, if there are $n$ trajectories and $k$ clusters, worst case complexity of the clustering becomes $\Theta(nk)$. Under normal circumstances, $k$ will be much smaller than $n$, hence the complexity can be approximated by $\Theta(n)$.
\subsection{Evaluation and Comparisons}
\label{chap4_sec:Model Evaluation}
\begin{figure}[!h]
  \centering
 \subfigure[Proposed method]{\includegraphics[width=0.158\textwidth]{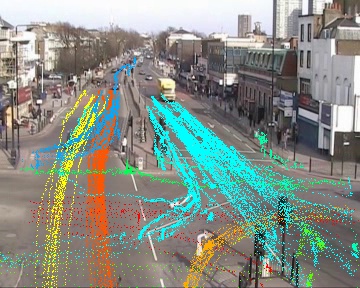}}
 \subfigure[DBSCAN]{\includegraphics[width=0.158\textwidth]{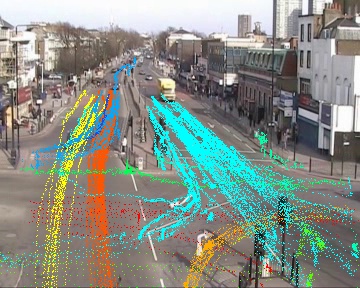}}                   
 \subfigure[mean shift]{\includegraphics[width=0.158\textwidth]{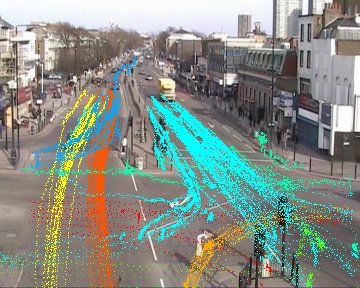}}
 \subfigure[TIGM - cluster 10]{\includegraphics[width=0.158\textwidth]{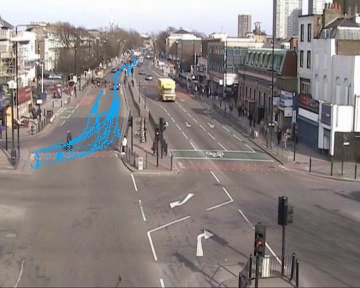}}                   
 \subfigure[DBSCAN - track 164]{\includegraphics[width=0.158\textwidth]{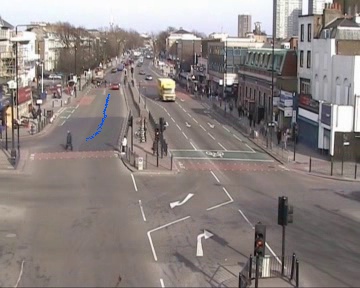}}
 \subfigure[mean shift - cluster 8]{\includegraphics[width=0.158\textwidth]{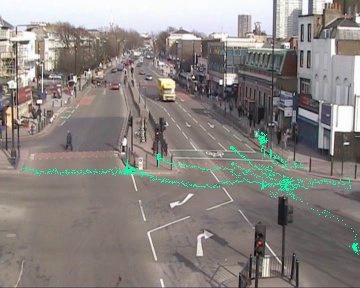}}
 \subfigure[TIGM - cluster 9]{\includegraphics[width=0.158\textwidth]{images/TRACK/COMPARISON/DBSCAN9.jpg}}   
 \subfigure[TIGM - track 89]{\includegraphics[width=0.158\textwidth]{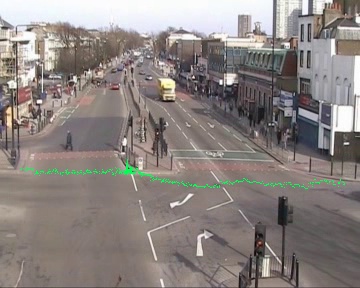}}                   
 \subfigure[TIGM - track 99]{\includegraphics[width=0.158\textwidth]{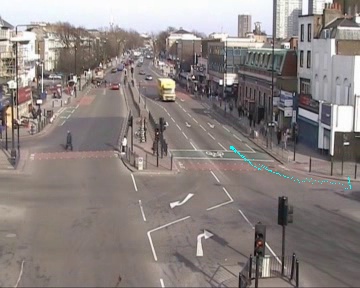}}
  \caption{Comparison of start point-based trajectory clustering based on QML dataset using TIGM ($\beta = 50$) with other techniques, namely mean shift ($quantile = 0.13$) and DBSCAN ($\epsilon = 30$, minimum samples = 1). (a - c) Clusters shown in unique colors, using different techniques. (d) Tracks of cluster 10 from TIGM (e) Track 164 from DBSCAN. (f) Tracks of cluster 8 from mean shift. (g) Tracks of cluster 10 from TIGM. (h) Track 89 from DBSCAN. (i) Track 99 from TIGM.}
  \label{chap4_Fig:Comparison}
\end{figure}

Next, the clustering results using two unsupervised and nonparametric approaches, namely mean shift~\cite{comaniciu2002mean} and DBSCAN~\cite{ester1996density} have been compared. Clustering has been applied on QML dataset based on start point. The results are shown in Fig.~\ref{chap4_Fig:Comparison}. Proposed method outperforms both approaches in the majority of the cases. Aforementioned techniques fail to group all trajectories into the right cluster. For example, as depicted in Fig.~\ref{chap4_Fig:Comparison}, track $164$ is getting associated with cluster 10. The proposed method resolves such cases. However, it cannot be claimed to be incorrect clustering (done by DBSCAN) since the criteria are different. Similarly, in mean shift clustering, tracks $89$ and $99$ are included in the  cluster $8$. Even though track $89$ can be logically grouped to cluster $8$, in TIGM and DBSCAN, it forms a new cluster. However, it may be incorrect to group track $99$ with cluster $8$. Ideally, it can be part of the cluster $9$, as done using TIGM and DBSCAN.

\begin{figure*}[!h]
  \centering
  \begin{minipage}[b]{0.24\textwidth}
 {\includegraphics[scale=0.24]{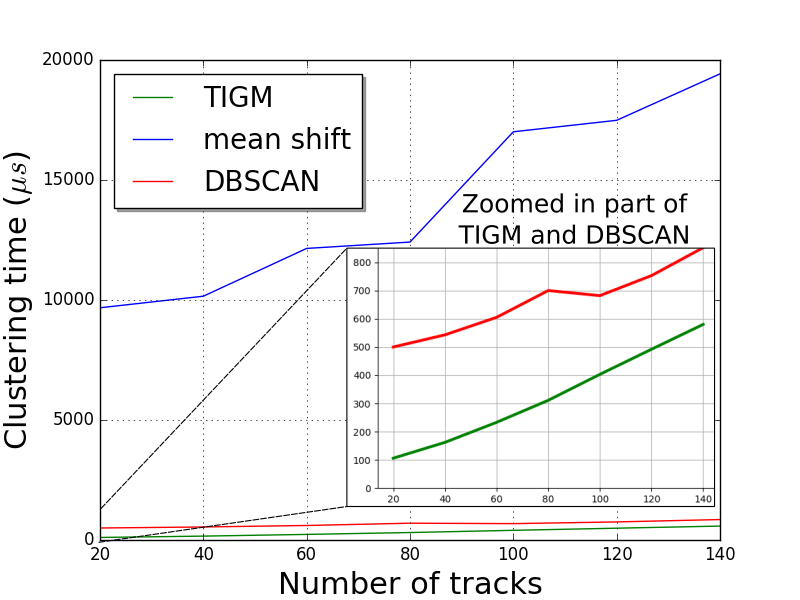}}  
  \caption{Plots showing the relationship of clustering time against the number of observations based on QML dataset considering only start point. Plot clearly shows the linear performance of TIGM as compared to other methods.} 
  \label{chap4_Fig:Performance_QML}   
   \end{minipage} 
   \hfill    
  \begin{minipage}[b]{0.24\textwidth}
  \centering
      \includegraphics[scale=0.24]{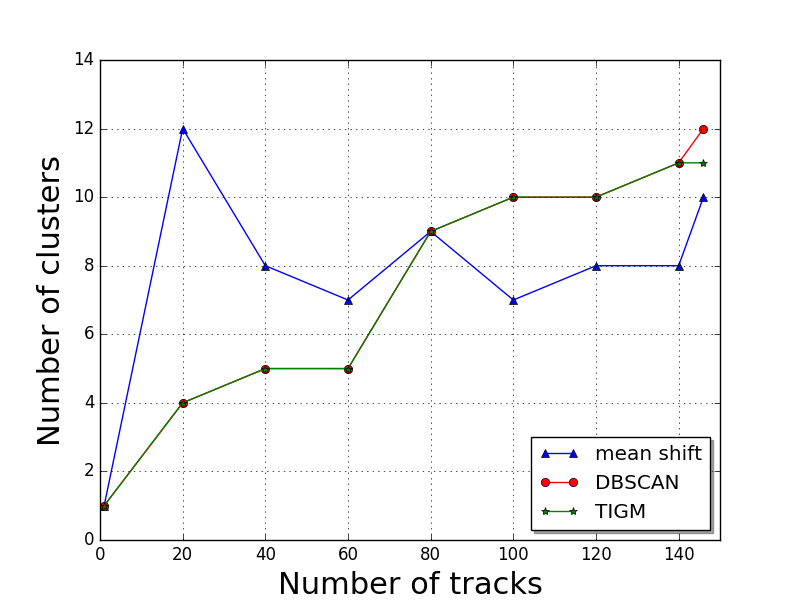}
    \caption{Plots comparing the relationship between number of trajectories and number of clusters formed for the three unsupervised nonparametric methods (mean shift, DBSCAN, and TIGM). Plot of TIGM is overlapping with DBSCAN at most of the points.}
  \label{chap4_Fig:ClusterFormationComp}  
   \end{minipage}  
   \hfill
   \begin{minipage}[b]{0.24\textwidth}
      \includegraphics[scale=0.24]{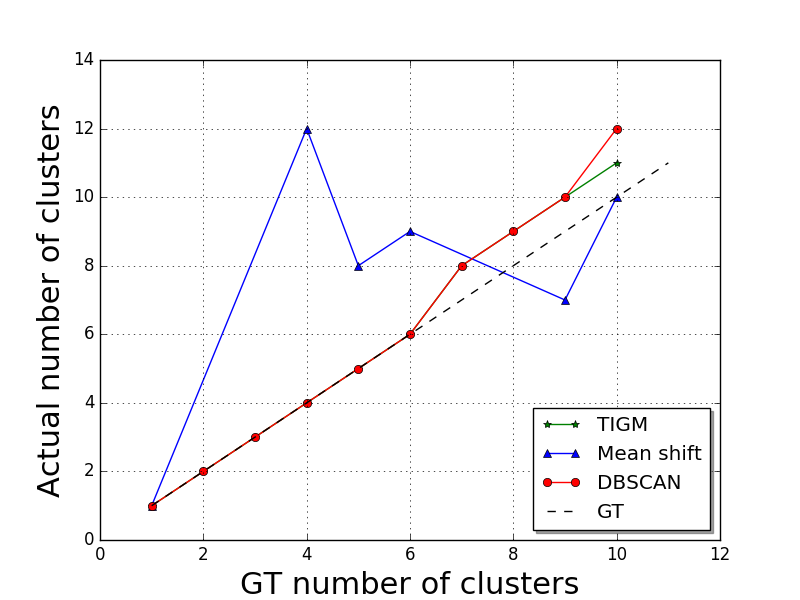}
    \caption{Ground truth (GT) vs. Actual number of clusters. Though the number of obtained clusters are different from ground truth as more trajectories arrive, all methods are able to come closer to the ground truth. TIGM and DBSCAN are able to learn them even with less number of trajectories.}
  \label{chap4_Fig:GTvsAct}  
  \end{minipage}
    \hfill
    \begin{minipage}[b]{0.24\textwidth}    
      \includegraphics[scale=0.24]{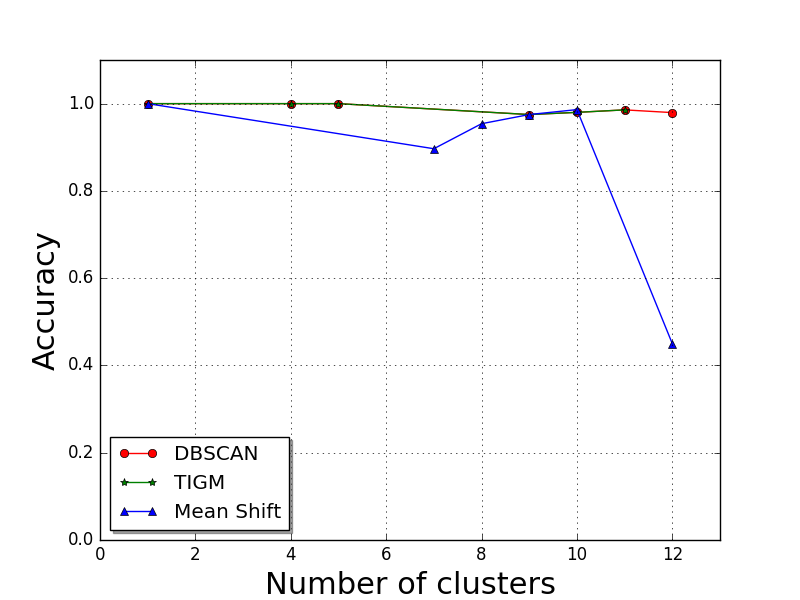}
    \caption{Accuracy of clustering plotted for different number of trajectories for QML dataset based on start point. Accuracy of DBSCAN and TIGM are comparable. However, mean shift gives varying accuracy.}
  \label{chap4_Fig:Learning_Accuracy}      
    \end{minipage}    
\end{figure*}
\begin{figure*}[!h]
  \centering
  \begin{minipage}[b]{0.39\textwidth}
        \includegraphics[width=1.0\textwidth]{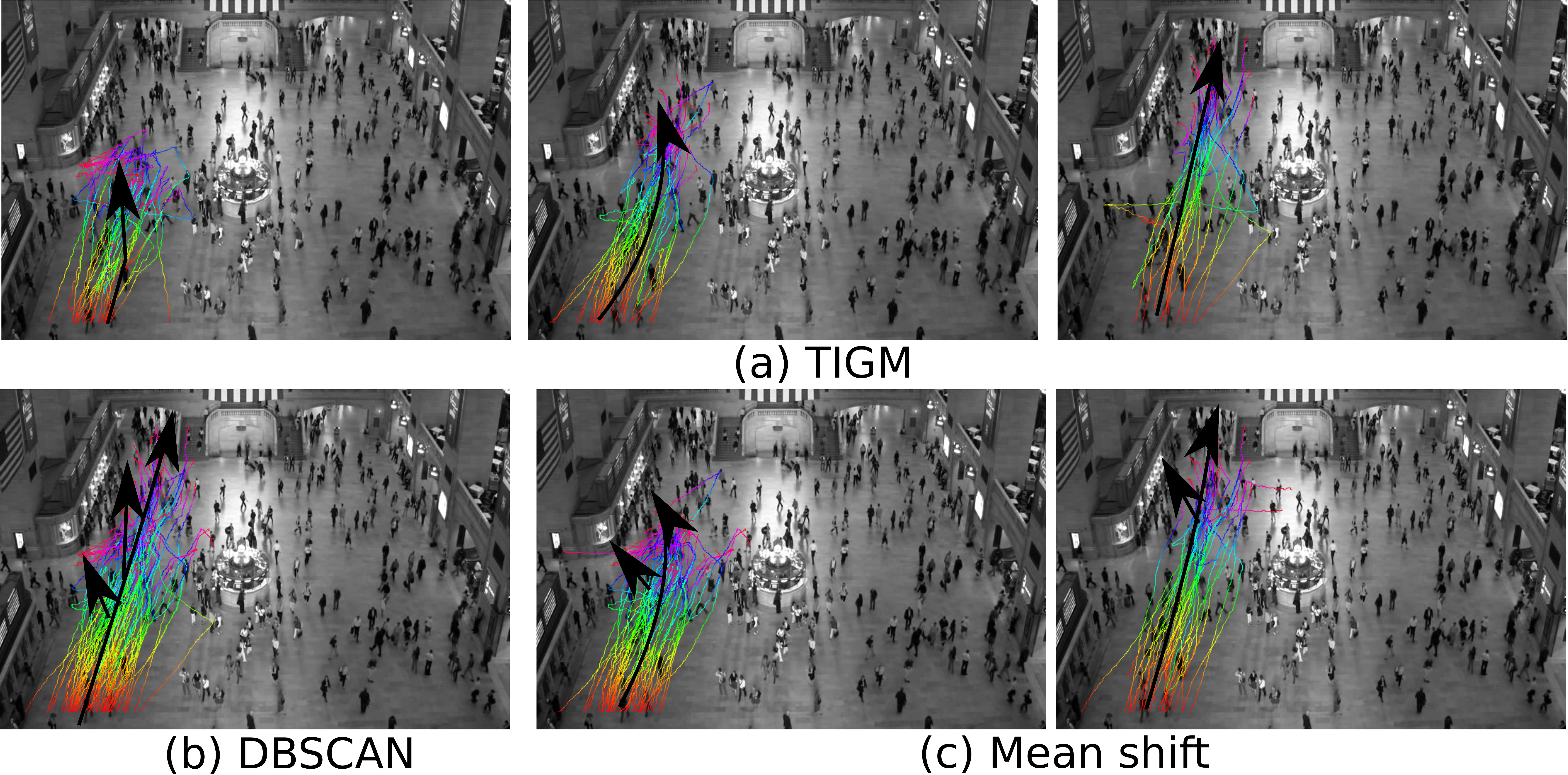}
    \caption{Depiction of cluster merging in DBSCAN and mean shift as compared to TIGM. (a) Three meaningful clusters using TIGM. (b) High concentration of trajectories in a path tends to merge the trajectories into a single cluster. (c) Mean shift also does similar cluster merging.}
  \label{chap4_Fig:Cluster_merging} 
  \end{minipage}   
      \hfill
\begin{minipage}[b]{0.29\textwidth}
      \includegraphics[width=1.0\textwidth]{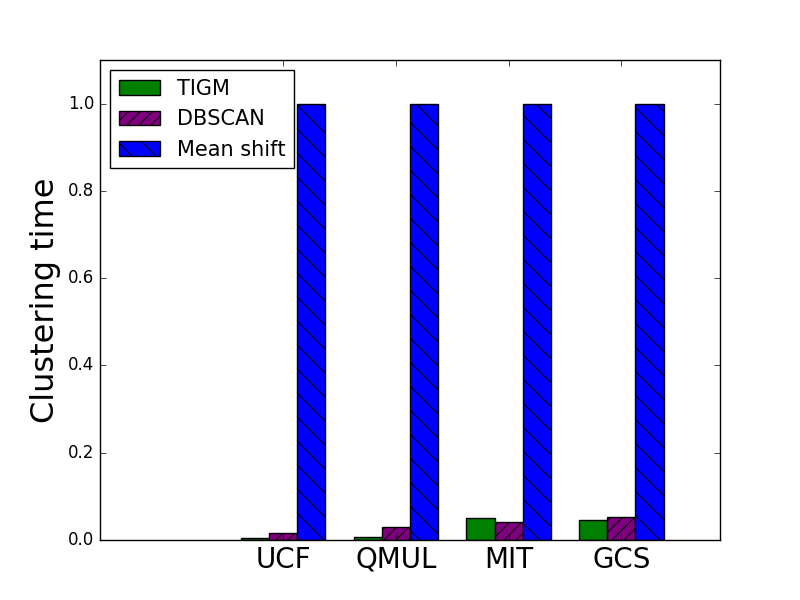}
    \caption{Depiction of clustering time across all
the datasets on a normalized scale. It can be observed that except in MIT dataset, TIGM performs better as compared to other methods.}
  \label{chap4_Fig:Performance_alldatasets} 
  \end{minipage}
      \hfill 
  \begin{minipage}[b]{0.29\textwidth}
      \includegraphics[width=1.0\textwidth]{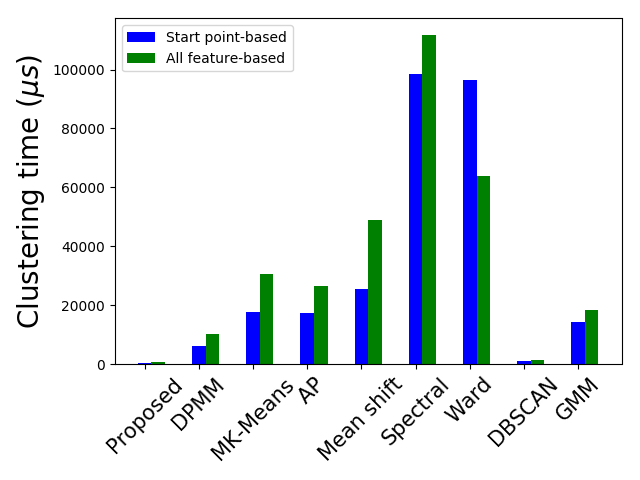}
    \caption{Depiction of clustering time for QML trajectories with state-of-the-art clustering algorithms. TIGM outperforms the state-of-the-art  algorithms. The closest competitor is DBSCAN.}
  \label{chap4_Fig:Performance}    
  \end{minipage}   
\end{figure*}
\begin{table}
\caption {Comparison among different datasets (A: Accuracy, P: Precision, R: Recall, $ms$: Minimum samples, $q$: Quantile). Note: normalized features have been used used for DBSCAN and mean shift clustering} 
\label{chap4_Tab:Accuracy} 
\begin{tabular}{>{$}l<{$} p{3.5cm}|p{0.8cm}|p{0.8cm}|p{0.8cm}} \hline  
\multicolumn{1}{l}{\text{Dataset}} &
\multicolumn{4}{c}{Parameter} \\  \cline{2-5}                           
& \text{Method} & \text{A (\%)} & \text{P (\%)} & \text{R (\%)}\\ \hline  \hline 
UCF & TIGM ($\beta=240$)\newline
		DBSCAN ($\epsilon=.83$, $ms=1$)\newline
		mean shift ($q = 0.07$)& 97.37\newline
				98.68\newline
				97.37	& 98.67\newline
						100\newline
						100 	& 98.67\newline
								98.68\newline
								97.37 \\ \hline  
QML & TIGM ($\beta=170$)\newline
		DBSCAN ($\epsilon=.83$, $ms=1$)\newline
		mean shift ($q = 0.0459$)& 92.47\newline
				91.10\newline
				89.73 	& 95.74\newline
						97.08\newline
						97.04 	& 96.43\newline
								93.66\newline
								92.25  \\ \hline 
MIT & TIGM ($\beta=50$)\newline
		DBSCAN ($\epsilon=.27$, $ms=1$)\newline
		mean shift($q = 0.01$)& 90.00\newline
				91.28\newline
				80.28	& 96.43\newline
						94.18\newline
						94.31 	& 93.10\newline
								96.74\newline
								84.36  \\ \hline 
GCS & TIGM ($\beta=120$)\newline
		DBSCAN ($\epsilon=.31$, $ms=1$)\newline
		mean shift($q = 0.01$)& 84.80\newline
				47.20\newline
				67.80 	& 94.43\newline
						80.55\newline
						85.39 	& 89.26\newline
								53.27\newline
								76.70  \\ \hline 
\end{tabular}
\end{table} 
     In order to evaluate these methods quantitatively, experiments have been conducted to calculate learning accuracy, precision, and recall. Firstly, the logical grouping of clusters has been done by an unbiased evaluator to prepare ground truths. Ground truths correspond to unique trajectory patterns that are possible in a typical traffic scene. For example, ground truth represents the trajectory patterns on a typical road. In GCS, they are the common movement patterns between a pair of source and destination. Wrong inclusion of a trajectory in the predicted group is categorized as False Positive (FP), and wrong exclusion from a coherent group is categorized as False Negative (FN). Rest of the trajectories are either True Positives (TP) or True Negatives (TN). Analysis of various clustering techniques including the proposed TIGM on QML dataset using start point is depicted in Figs.~\ref{chap4_Fig:Performance_QML} - \ref{chap4_Fig:Learning_Accuracy}. Fig.~\ref{chap4_Fig:Performance_QML} reveals that TIGM takes linear time. The plots are shown to demonstrate the predictability of the model. It may be observed that DBSCAN matches with TIGM in cluster prediction with respect to ground truth and their accuracies. Though the cluster prediction of mean shift gets better with the increase in the number of tracks, accuracy has been found to be varying.

Table \ref{chap4_Tab:Accuracy} summarizes the experimental results in terms of accuracy, precision, and recall. For UCF ($76$ trajectories) and QML ($146$ trajectories), all five dimensions are used. In MIT ($449$ trajectories) and GCS (initial $1000$ trajectories) datasets, duration ($t_{dur}$) has been omitted since it has been found to be difficult to clearly discriminate between shorter and longer trajectories with the same start and end positions. The parameter values have been adjusted to create common patterns corresponding to source-destination pairs such that the methods can be compared across various datasets. It may be observed that the performance of TIGM is similar to DBSCAN when applied on road trajectories. However, TIGM performs better than DBSCAN and mean shift on GCS data. From the analysis, it has been inferred that DBSCAN performs poorly due to the spread of starting and ending points in the scene. When the trajectories are dense around a particular locality, DBSCAN typically groups them to form bigger clusters, as illustrated in Fig.~\ref{chap4_Fig:Cluster_merging}. By lowering $\epsilon$, such effect can be avoided. However, this leads to a higher number of cluster formation, causing accuracy reduction due to higher false negatives. The clustering time has been compared with DBSCAN and mean shift across all datasets and the results are shown in Fig.\ref{chap4_Fig:Performance_alldatasets}.

In addition to nonparametric clustering algorithms, comparisons with the performance of other unsupervised methods, namely Minibatch K-means (MK-means)~\cite{sculley2010web}, Ward~\cite{WardAgglo}, Affinity propagation (AP)~\cite{Affinity2007clustering}, Spectral~\cite{Spectralvon2007tutorial}, Gaussian Mixture Model (GMM)~\cite{EM_GMM}  have also been done. TIGM has been found to be reasonably faster than other unsupervised clustering methods as shown in Fig.~\ref{chap4_Fig:Performance}. 
  
\subsection{Summary of Results}
\label{chap4_sec:Result_Discussion} 
    Trajectory-based learning is one of the common ways to analyze and interpret activities in a scene. From the experiments, it has been observed that the proposed method is able to learn unique trajectory patterns in a traffic scene. Using a semi-supervised approach (by pre-initializing normal clusters with representative trajectories), the learned patterns can be identified as normal or abnormal in terms of posterior probabilities represented in (\ref{chap4_equation:INF5}).  For a new trajectory, classification and behavior analysis can be done using the learned model. With more specific features such as velocity or curvature, activity detection and classification can be done. Statistics corresponding to unique trajectory patterns can be used by traffic authorities for better planning of road networks, safety analysis, risk assessment, etc. The temporal extension of TIGM, i.e., the DEM reflects the traffic characteristics over time. This is useful in traffic flow analysis, congestion study, traffic density analysis, adaptive traffic signal scheduling, traffic re-routing strategies since the cluster dynamics provide valuable traffic information about the paths, entry,  and exit points. 

\section{Discussion}
\label{chap4_sec:Discussion}
\subsection{TIGM-DEM Vs. Relevant Methods}
 Despite producing strikingly similar results in terms of coherent groups, proposed TIGM has other advantages over existing methods. The proposed method clusters the trajectories in $\Theta(nk)$ time, while mean shift takes $O(k\text{ }n\text{ }log(n))$ time and for higher dimensional data, the time complexity can be as large as $O(k\text{ }n^2)$. Here, $n$ is the number of observations and $k$ the number of clusters. DBSCAN does clustering in $O(n\text{ }log(n))$ time in an average case and $O(n^2)$ in the worst case.  DBSCAN and mean shift algorithms are statistical in nature, i.e., they consider all data points together. Such methods are useful for data mining related analysis. Though the proposed method can be used by going for model parameter convergence, temporal information may be lost in the clustering process. DBSCAN and mean shift have significance in clustering in a wide variety of applications. In traffic monitoring applications, trajectories arrive in temporal order, one at a time, making the incremental model ideal to represent the scene dynamics. Moreover, by introducing the parameter $\beta$ (concentration radius) as a distance, it is intuitive to estimate the parameter unlike DBSCAN or mean shift. Also, the proposed clustering starts with no prior (in the inference). The prior is built incrementally during the clustering process. Inference mechanism allows unassigning any observation without affecting the cluster labels. This is the key to build the DEM that can predict/describe status of the scene across different time frames. The model reflecting the changing dynamics is a key feature of the proposed method. 
This means, abnormal activity detection can be more adaptive rather than following a strict threshold in traffic monitoring applications. Moreover, there is no need to map the cluster labels as they are maintained in the temporal plane.

   Like Dynamic-Dual-HDP\cite{wang2011trajectory} approach, the proposed DEM is able to model a scene dynamically, thus catering to learning the scene parameters dynamically. Moreover, only essential information needs to be stored using the proposed method, rather than storing a complete trajectory. However, Dynamic-Dual-HDP looks for convergence, thus may lose some important temporal characteristics. Unlike \cite{zhu2005time}, the temporal aspects are handled inherently by using observations in the temporal order in the proposed DEM framework. In the work \cite{IncrementalDPMM},  though a new model based on DFT features has been introduced, it does not explore much on trajectories from different contexts. Moreover, the proposed method has the capability to handle scene dynamics with a simple extension of TIGM. It is able to produce meaningful semantic regions even on challenging dataset such  as GCS in comparisons with \cite{xu2015unsupervised}.  
 
Deep learning approches are mainly used in object detection and prediction related tasks for road traffic applications. The method proposed in~\cite{2017_Zhang_Traffic_Monitoring_DNN} uses deep learning for object detection. It uses a heuristic approach for traffic flow estimation in highways. However, DEM can capture the flow information in terms of model parameters. In~\cite{Deep_trafficflow_2015_Loop, zhao2017lstm, 2016_Speed_deep} traffic prediction has been done using DNNs trained with historical data at a macro level based on sensors spread over a larger geography. However, these studies lack the essential visual information such as object and their interactions. Though DEM cannot predict the traffic flow, it can act as a building block in the design of traffic monitoring applications suitable for congestion control, future infrastructure planning, and anomaly detection. DEM experiments have been conducted using only three features, however being a generic model, it can produce better results with automatically extracted features using DNN frameworks.
\subsection{Limitations}
For the simplicity of modeling, a single parameter ($\beta$) has been assumed for the entire clustering. However, it may so happen that  different clusters follow distributions of varying spread. In such cases, the likelihood functions need to be learned to avoid observations going to wrong clusters, even though they are spatially apart. This can happen to observations belonging to less frequent patterns with their mean being closer to a cluster of higher spread. The proposed method cannot be used in situations where trajectory patterns are random in nature, as the accuracy may be affected due to the approximation of trajectories to start and end positions as well as  the duration. By incorporating more dimensions like curvature, slope or direction, better accuracy can be achieved. The proposed model allows using other suitable distance measures as well.  If noisy trajectories such as truncated tracks are present, the clusters formed may not truly represent the scene since this method heavily relies on the accuracy of tracking algorithms. Such noisy trajectories typically form new clusters as their characteristics (or features) are different from the normal ones. With the advances in machine learning, better tracking algorithms are available, thus such situations can be avoided.

\begin{figure}[!h]
  \centering
      \includegraphics[scale = 0.37]{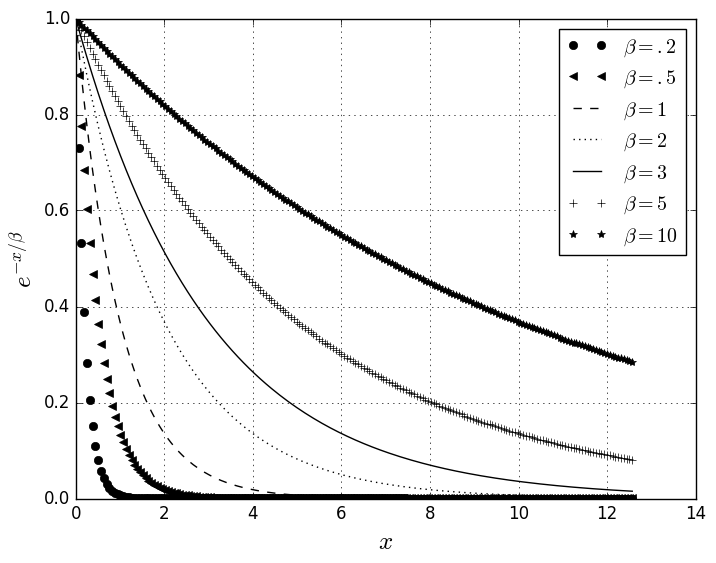}
    \caption{Depiction of decay function behavior with different values of $\beta$. Different likelihood functions  correspond to clusters of different spreads.}
  \label{chap4_Fig:BetaEDecay}  
\end{figure}
\section{Conclusion}
\label{chap4_sec:conclusion}
This paper introduces a trajectory model with the help of DPMM from a new perspective by using a distance measure and temporal correlation of the observations. The model has been extended temporally to consider the dynamic behavior of objects in a scene to reflect the traffic variations. An incremental approach has been used to build clusters without any prior knowledge. The model has been validated on a wide variety of video datasets. The proposed model is able to cluster trajectories meaningfully with higher accuracy even with lesser computation time. The proposed model can be used for building real-time traffic analysis frameworks since it can learn and refine frequently occurring patterns in an unsupervised and nonparametric way. Though the main objective of the proposed model is for visual surveillance applications, it may also be used in applications requiring representation of changing scene dynamics involving observations of temporal dependencies. 

There are rooms for improvement at various levels. Firstly, likelihoods corresponding to distributions of varying spreads need to be learned. Fig.~\ref{chap4_Fig:BetaEDecay} gives a clue about how different spreads can be handled. With one round of Gibbs sampling, $\beta$ of the respective cluster can be learned. This acts as a clue to learn the likelihood for each cluster. Secondly, Mahanolabis distance can be used for better approximation of elliptical distributions. Lastly, application for real-time monitoring of road traffic can be built.


%

%
%
%
%
%

\ifCLASSOPTIONcaptionsoff
  \newpage
\fi



%
%
%

\bibliographystyle{plain}


%
\begin{IEEEbiography}[{\includegraphics[width=1in,height=1.25in,clip,keepaspectratio]{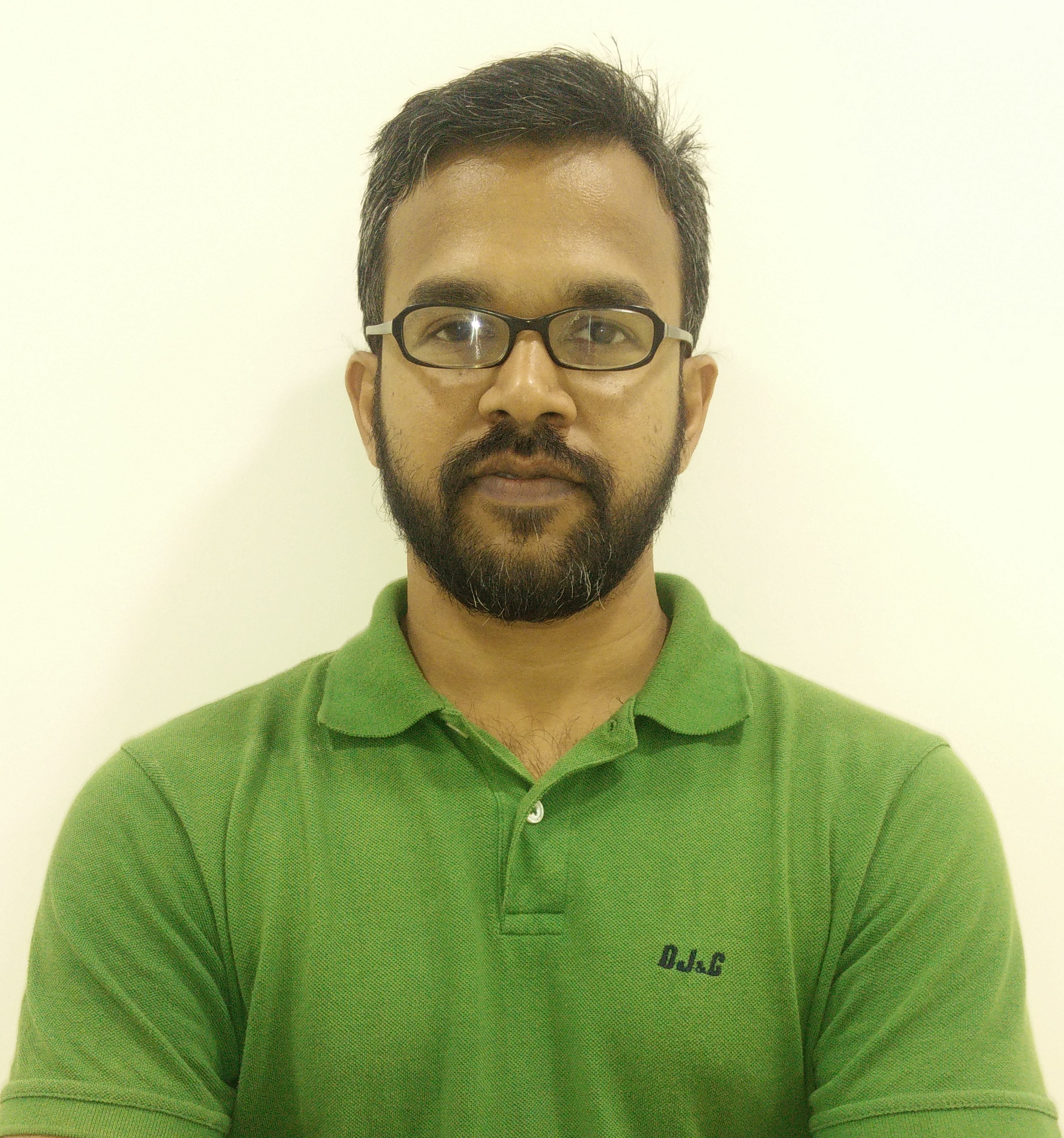}}]{Kelathodi Kumaran Santhosh} is a research scholar in the School of Electrical Sciences, IIT Bhubaneswar, India. He joined a Ph.D. program for resuming his research work that can help humanity. His interests are in the development of vision based applications that can replace human factor. He is a member of IEEE. Prior to joining IIT Bhubaneswar, he worked for Huawei Technologies India Pvt. Ltd. for 10 years (2005-2015) and in Defence Research Development Organization (DRDO) as a Scientist for around 2 years (2003-2004). During his tenure with Huawei, he has worked in many signalling protocols such as Diameter, Radius, SIP etc. in the role of a developer, technical leader, project manager and also served the product lines HSS, CSCF etc. in Huawei China as a support engineer. In DRDO, he worked on target tracking algorithms on radar data.  More information on Santhosh can be found at https://sites.google.com/site/santhoshkelathodi.
\end{IEEEbiography}
\vspace{-1cm}
\begin{IEEEbiography}[{\includegraphics[width=1in,height=1.25in,clip,keepaspectratio]{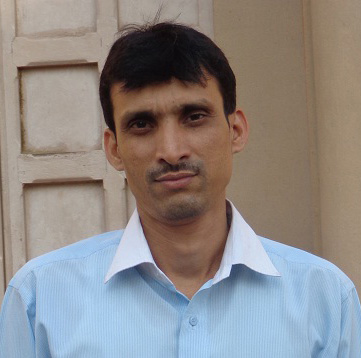}}]{Dr. Debi Prosad Dogra} is an Assistant Professor in the School of Electrical Sciences, IIT Bhubaneswar, India. He received his M.Tech degree from IIT Kanpur in 2003 after completing his B.Tech. (2001) from HIT Haldia, India. After finishing his masters, he joined Haldia Institute of Technology as a faculty members in the Department of Computer Sc. \& Engineering (2003-2006). He has worked with ETRI, South Korea during 2006-2007 as a researcher. Dr. Dogra has published more than 75 international journal and conference papers in the areas of computer vision, image segmentation, and healthcare analysis.  He is a member of IEEE. More information on Dr. Dogra can be found at \url{http://www.iitbbs.ac.in/profile.php/dpdogra}.
\end{IEEEbiography}
\vspace{-1cm}
\begin{IEEEbiography}[{\includegraphics[width=1in,height=1.25in,clip,keepaspectratio]{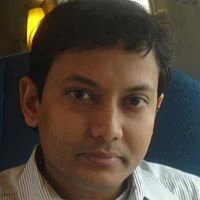}}]{Dr. Partha Pratim Roy} has obtained his M.S. and Ph. D. degrees in the year of 2006 and 2010, respectively at Autonomous University of Barcelona, Spainis. Presently he is an Assistant Professor in the Department of Computer Science and Engineering, IIT Roorkee, India in 2014. Prior to joining, IIT Roorkee, Dr. Roy was with Advanced Technology Group, Samsung Research Institute Noida, India during 2013-2014. Dr. Roy was with Synchromedia Lab, Canada in 2013 and RFAI Lab, France in 2012 as postdoctoral research fellow. His research interests are Pattern Recognition, Multilingual Text Recognition, Biometrics, Computer Vision, Image Segmentation, Machine Learning, and Sequence Classification. He has published more than 100 papers in international journals and conferences. 
\end{IEEEbiography}
\vspace{-1cm}
\begin{IEEEbiography}[{\includegraphics[width=1in,height=1.25in,clip,keepaspectratio]{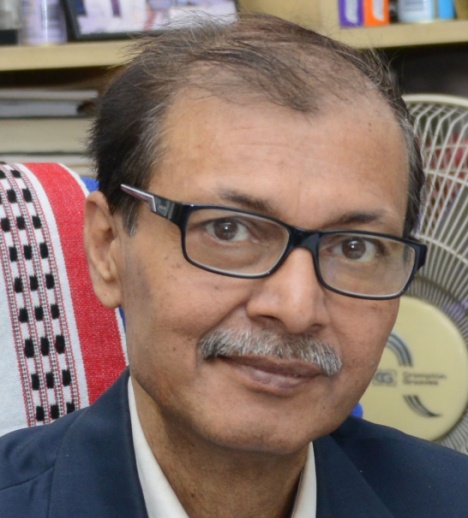}}]{Prof. Bidyut Baran Chaudhuri} retired as Head, Computer Vision and Pattern Recognition Unit, Indian Statistical Institute in December, 2015 and then joined as INAE Distinguished Professor and J.C. Bose Fellow in the same institute to work till December, 2018. Thereafter he joined Techno India University as Pro Vice-Chancellor (Academic) which he is continuing. He did his B.Tech. and M.Tech from Calcutta University and Ph.D. from IIT, Kanpur in 1972, 1974 and 1980, respectively. During 1981-82 he did his Post doc in Queen’s University, U.K. as Leverhulme Overseas Fellow. He also worked as a Visiting Faculty at Tech. University, Hannover during 1987-88 as well as GSF (now Leibnitz Institute) during 1985, 1990 and 1992. He joined as regular faculty in Indian Statistical Institute in 1978. During his tenure, he acted as UNDP KBCS project coordinator in 1992-94, Jawaharlal Nehru fellow 2002-04 and did held several other honorable positions. His research interests include Pattern Recognition, Image Processing, Computer Vision, Speech Language Processing, OCR, Machine learning etc. and published 425 research papers and 6 technical books. Prof. Chaudhuri is an Associate Editor of International Journal of Document Analysis and Recognition (IJDAR) and Int. Journal of Pattern Recognition and Artificial Intelligence (IJPRI). Earlier, he served as Associate Editor of Pattern Recognition, Pattern Recognition Letters etc. He is a fellow of IEEE, International Association for Pattern Recognition (IAPR), Third World Academy of Science (TWAS), Indian National Science Academy (INSA), National Academy of Science (NASc), Indian National Academy of Engineering (IETE), West Bengal Academy of Science \& Technology, Optical Society of India etc. 
\end{IEEEbiography}



\end{document}